\documentclass[journal,comsoc]{IEEEtran}
\usepackage{epsfig}
\usepackage{epstopdf}
\usepackage{subfigure} 
\usepackage{amsmath}
\usepackage{graphicx}
\usepackage{mathrsfs}
\usepackage{color}
\usepackage[T1]{fontenc}
\usepackage{cite}
\usepackage[linesnumbered,ruled]{algorithm2e}
\usepackage{multirow}
\usepackage{makecell}
\usepackage[cmintegrals]{newtxmath}
\usepackage{url}
\DeclareGraphicsExtensions{.pdf,.png,.jpg,.fig}
\SetKwRepeat{Do}{do}{while}
\DeclareMathOperator*{\argminA}{arg\,min}

\newcommand\norm[1]{\left\lVert#1\right\rVert}
\newcommand{\RN}[1]{\textup{\uppercase\expandafter{\romannumeral#1}}}
\begin{document}
\title{Pothole Detection Based on Disparity Transformation and Road Surface Modeling}
\author{Rui~Fan,~\IEEEmembership{Member,~IEEE},
Umar~Ozgunalp,~\IEEEmembership{Member,~IEEE},
Brett~Hosking,~\IEEEmembership{Member,~IEEE},
\\
Ming~Liu,~\IEEEmembership{Senior Member,~IEEE},~Ioannis~Pitas,~\IEEEmembership{Fellow,~IEEE}  
\thanks{R. Fan is with the Robotics Institute, Department of Electronic and Computer Engineering, the Hong Kong University of Science and Technology, Hong Kong SAR, China (e-mail: rui.fan@ieee.org).}
\thanks{U. Ozgunalp is with the Electrical and Electronic Engineering, Near East University, Nicosia, Cyprus (e-mail: umar.ozgunalp@neu.edu.tr).}
\thanks{B. Hosking is with the National Oceanography Centre, Southampton, U.K. (e-mail: wilski@noc.ac.uk).}
\thanks{M. Liu is with the Robotics Institute, Department of Electronic and Computer Engineering, the Hong Kong University of Science and Technology, Hong Kong SAR, China (e-mail: eelium@ust.hk).}
\thanks{I. Pitas is with the School of Informatics, Aristotle University of Thessaloniki, Thessaloniki, Greece (e-mail: pitas@aiia.csd.auth.gr).}
\thanks{This work was supported by the National Natural Science Foundation of China, under grant No. U1713211, the Research Grant Council of Hong Kong SAR Government, China, under Projects No. 11210017 and No. 21202816, and the Shenzhen Science, Technology and Innovation Commission (SZSTI), under grant JCYJ20160428154842603, awarded to Prof. Ming Liu. }
}

\markboth{IEEE Transactions on Image Processing}
{Fan \MakeLowercase{\textit{et al.}}: Pothole Detection Based on Disparity Transformation and Road Surface Modeling}
\maketitle
\begin{abstract}
Pothole detection is one of the most important tasks for road maintenance. Computer vision approaches are generally based on either 2D road image analysis or 3D road surface modeling. However, these two categories are always used independently. Furthermore, the pothole detection accuracy is still far from satisfactory. Therefore, in this paper, we present a robust pothole detection algorithm that is both accurate and computationally efficient. A dense disparity map is first transformed to better distinguish between damaged and undamaged road areas. To achieve greater  disparity transformation efficiency, golden section search and dynamic programming are utilized to estimate the transformation parameters. Otsu's thresholding method is then used to extract potential undamaged road areas from the transformed disparity map. The disparities in the extracted areas are modeled by a quadratic surface using least squares fitting. To improve disparity map modeling robustness, the surface normal is also integrated into the surface modeling process. Furthermore, random sample consensus is utilized to reduce the effects caused by outliers. By comparing the difference between the actual and modeled disparity maps, the potholes can be detected accurately. Finally, the point clouds of the detected potholes are extracted from the reconstructed 3D road surface. The experimental results show that the successful detection accuracy of the proposed system is around $\boldsymbol{98.7\%}$ and the overall pixel-level accuracy is approximately $\boldsymbol{99.6\%}$. 
\end{abstract}
\section{Introduction}
\label{sec.introduction}
\IEEEPARstart{R}OAD potholes are considerably large structural failures on the road surface. They are caused by contraction and expansion of the road surface as rainwater permeates into the ground \cite{Miller2014}. To ensure traffic safety, it is crucial and necessary to frequently inspect and repair road potholes \cite{Mathavan2015}.  Currently, potholes are regularly detected and reported by certified inspectors and structural engineers \cite{Kim2014}. This task is, however,  time-consuming and tedious \cite{fan2019real}. Furthermore, the detection results are always subjective, because they depend entirely on personnel experience \cite{Koch2015}. Therefore,  automated pothole detection systems have been developed to recognize and localize potholes both efficiently and objectively. 

Over the past decade, various technologies, such as active and passive sensing, have been utilized to acquire road data and aid personnel in detecting  road potholes \cite{Koch2015}. For example, Tsai and Chatterjee \cite{Tsai2017} mounted two laser scanners on a digital inspection vehicle (DIV) to collect 3D road surface data. These data were then processed using either semi or fully automatic methods for pothole detection. Such systems ensure personnel safety, but also reduce the need for manual intervention \cite{Tsai2017}. 
Furthermore, by comparing the road data collected over different periods, the traffic flow can be evaluated and the future road condition can be predicted \cite{Mathavan2015}. The remainder of this section presents the state-of-the-art pothole detection algorithms and highlights the motivation, contributions and outline of this paper. 
\subsection{State of the Art in Road Pothole Detection}
\label{sec.related_work}
\subsubsection{2D Image Analysis-Based Pothole Detection Algorithms}
\label{sec.2d_based_algorithms}
There are typically four main steps used in 2D image analysis-based pothole detection algorithms: a) image preprocessing; b) image segmentation; c) shape extraction; d) object recognition \cite{Koch2015}. A color or gray-scale road image is first preprocessed, e.g., using morphological filters \cite{Pitas2000}, to reduce image noise and enhance the pothole outline \cite{Koch2015, Li2016}. The preprocessed road image is then segmented using histogram-based thresholding methods, such as Otsu's \cite{Buza2013} or the triangle \cite{Koch2011} method.  Otsu's method minimizes the intra-class variance and performs better in terms of separating damaged and undamaged road areas \cite{Buza2013}.  The extracted region is then modeled by an ellipse \cite{Koch2011}. Finally, the image texture within the ellipse is compared with the undamaged road area texture. If the former is coarser than the latter, the ellipse is considered to be a pothole \cite{Koch2015}.

However, both color and gray-scale image segmentation techniques are severely affected by various factors, most notably by poor illumination conditions \cite{Jahanshahi2012}. Therefore, some authors proposed to perform segmentation on the depth maps, which has shown to achieve better performance when separating damaged and undamaged road areas \cite{Tsai2017, Jahanshahi2012}. Furthermore, the shapes of actual potholes are always irregular, making the geometric and textural assumptions occasionally unreliable. Moreover, the pothole's 3D spatial structure cannot always be explicitly illustrated in 2D road images \cite{Fan2018}. Therefore, 3D road surface information is required to measure pothole volumes. In general, 3D road surface modeling-based pothole detection algorithms are more than capable of overcoming the aforementioned disadvantages.
\subsubsection{3D Road Surface Modeling-Based Pothole Detection Algorithms}
\label{sec.3d_based_algorithms}
The 3D road data used for pothole detection is generally provided by laser scanners \cite{Tsai2017}, Microsoft Kinect sensors \cite{Jahanshahi2012}, or passive sensors \cite{Fan2018, Zhang2012, fan2018real, Zhang2014, Mikhailiuk2016, fan2019real}. Laser scanners mounted on DIVs are typically used for accurate  3D road surface reconstruction. However, the purchase of laser scanning equipment and its long-term maintenance are still very expensive \cite{Kim2014}. 
The Microsoft Kinect sensors were initially designed for indoor use. Therefore, they greatly suffer from infra-red saturation in direct sunlight \cite{Cruz2012a}. For this reason, passive sensors, such as a single movable camera or multiple synchronized cameras, are more suitable for acquiring 3D road data and for pothole  detection\cite{Fan2018, Hartley2003}. 
 For example, Zhang and Elaksher \cite{Zhang2012} mounted a single camera on an unmanned aerial vehicle (UAV) to reconstruct the road surface via Structure from Motion (SfM) \cite{Hartley2003}. A variety of stereo vision-based pothole detection methods have been developed as well \cite{Zhang2014, Mikhailiuk2016}.
 The 3D point cloud generated from a disparity map was interpolated into a quadratic surface using least squares fitting (LSF) \cite{Zhang2014}. The potholes were then recognized by comparing the difference between the 3D point cloud and the fitted quadratic surface. In \cite{Mikhailiuk2016},  the surface modeling was performed on disparity maps instead of the point clouds, and random sample consensus (RANSAC) \cite{fischler1981random} was used to improve the pothole detection robustness. 
\subsection{Motivation}
\label{sec.pd_motivation}
Currently, laser scanning is still the main technology used to provide 3D road information for pothole detection, while other technologies, such as passive sensing, are under-utilized \cite{Mathavan2015}. However, DIVs are not widely used, primarily because of the initial cost but also because their routine operation and long-term maintenance are still very costly \cite{Koch2011}. Therefore, the trend is to  equip DIVs with inexpensive, portable and durable sensors, such as digital cameras, for 3D road data acquisition. Stereo road image pairs can be used to calculate the disparity maps \cite{fan2018real}, which essentially represent the 3D road surface geometry. Recently, due to some major advances in computer stereo vision, road surface geometry can be reconstructed with a  three-millimeter accuracy \cite{Fan2018, fan2019real}. Additionally, stereo cameras used for road data acquisition are inexpensive, portable and adaptable for different DIV types.

So far, comprehensive studies have been made in both 2D image analysis-based and 3D road surface modeling-based pothole detection. Unfortunately, these algorithms are always used independently \cite{Koch2015}. Furthermore, pothole detection accuracy is still far from satisfactory \cite{Koch2015}. Exploring effective approaches for disparity map preprocessing, by applying 2D image processing algorithms, is therefore also a popular area of research that requires more attention. 
Only the disparities in the potential undamaged road areas are then used for disparity map modeling. 

Moreover, the surface normal vector is a very important descriptor, which is, however,  rarely utilized in existing 3D road surface modeling-based pothole detection algorithms. In this paper, we improve  disparity map modeling by eliminating the disparities whose surface normals differ significantly from the expected ones. 
\subsection{Novel Contributions}
\label{sec.contribution}
\begin{figure*}[!t]
	\begin{center}
		\centering
		\includegraphics[width=0.80\textwidth]{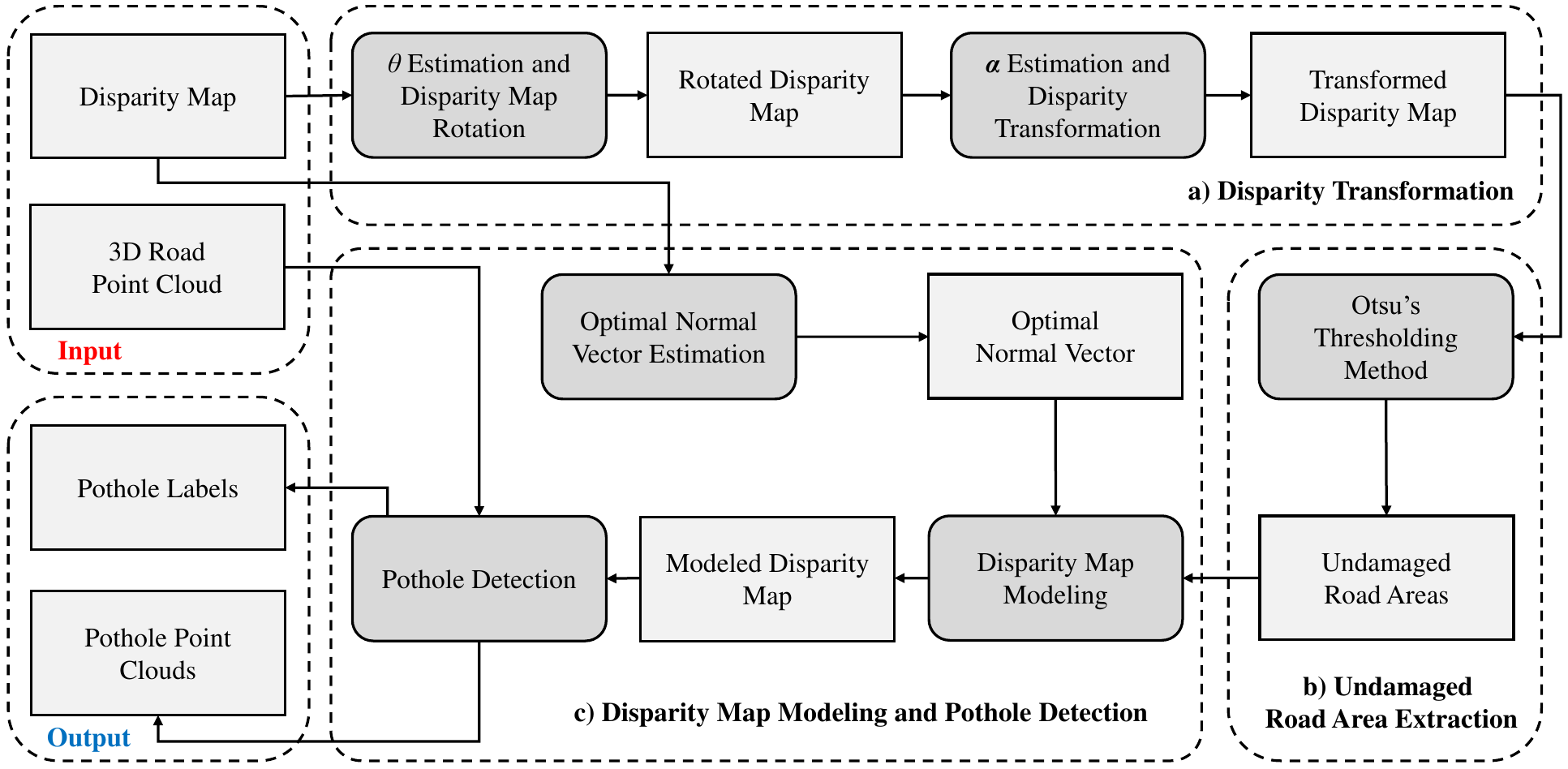}
		\centering
		\caption{The block diagram of the proposed pothole detection algorithm.}
		\label{fig.block_diagram}
	\end{center}
\end{figure*}
In this paper, a robust stereo vision-based pothole detection system is introduced. The main contributions are: a) a novel disparity transformation algorithm; b) a robust disparity map modeling algorithm; c) three pothole detection datasets which have been made publicly available for research purposes. These datasets are also used in our experiments for assessing pothole detection accuracy. 

 Since the disparities in damaged road areas can severely affect the accuracy of disparity modeling, we first transform the disparity maps to better distinguish between damaged and undamaged road areas. To achieve greater processing efficiency, we use golden section search (GSS) \cite{Pedregal2006} and dynamic programming (DP) \cite{Ozgunalp2017} to estimate the transformation parameters. Otsu's thresholding method \cite{Otsu1979} is then performed on the transformed disparity map to extract the undamaged road areas, where the disparities can be modeled by a quadratic surface using LSF. 
To improve the robustness of disparity map modeling, the surface normal information is also integrated into the modeling process. Furthermore, RANSAC is utilized to reduce the effects caused by any potential outliers. By comparing the difference between the actual and modeled disparity maps, the potholes can be detected effectively. Finally, different potholes are labeled using connected component labeling (CCL) \cite{dillencourt1992general} and their point clouds are extracted from the reconstructed 3D road surface. 
\subsection{Paper Outline}
\label{sec.outline}
The remainder of this paper is structured as follows: Section \ref{sec.algorithm_description} details the proposed pothole detection algorithm.  The experimental results for performance evaluation are illustrated in Section \ref{sec.experimental_results}. Finally, Section \ref{sec.conclusion} summarizes the paper and provides recommendations for future work. 
\begin{figure}[!t]
	\centering
		\subfigure[]{
		\includegraphics[width=0.43\textwidth]{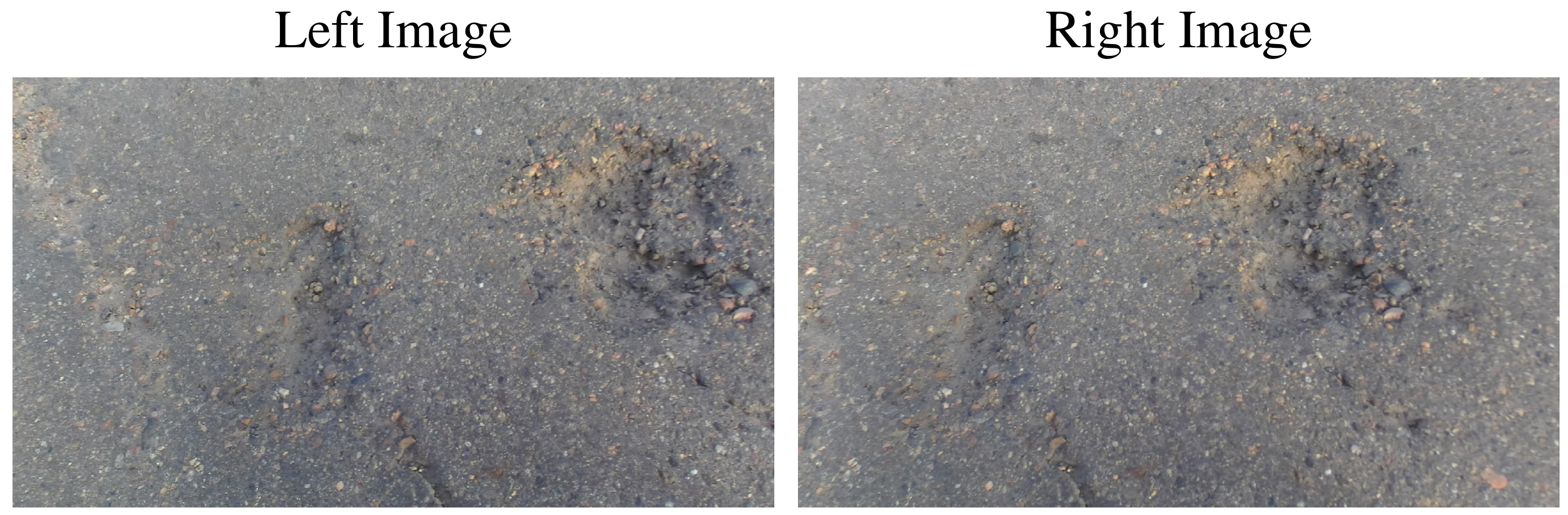}
		\label{fig.stereo_img}
	}
	\subfigure[]{
		\includegraphics[width=0.242\textwidth]{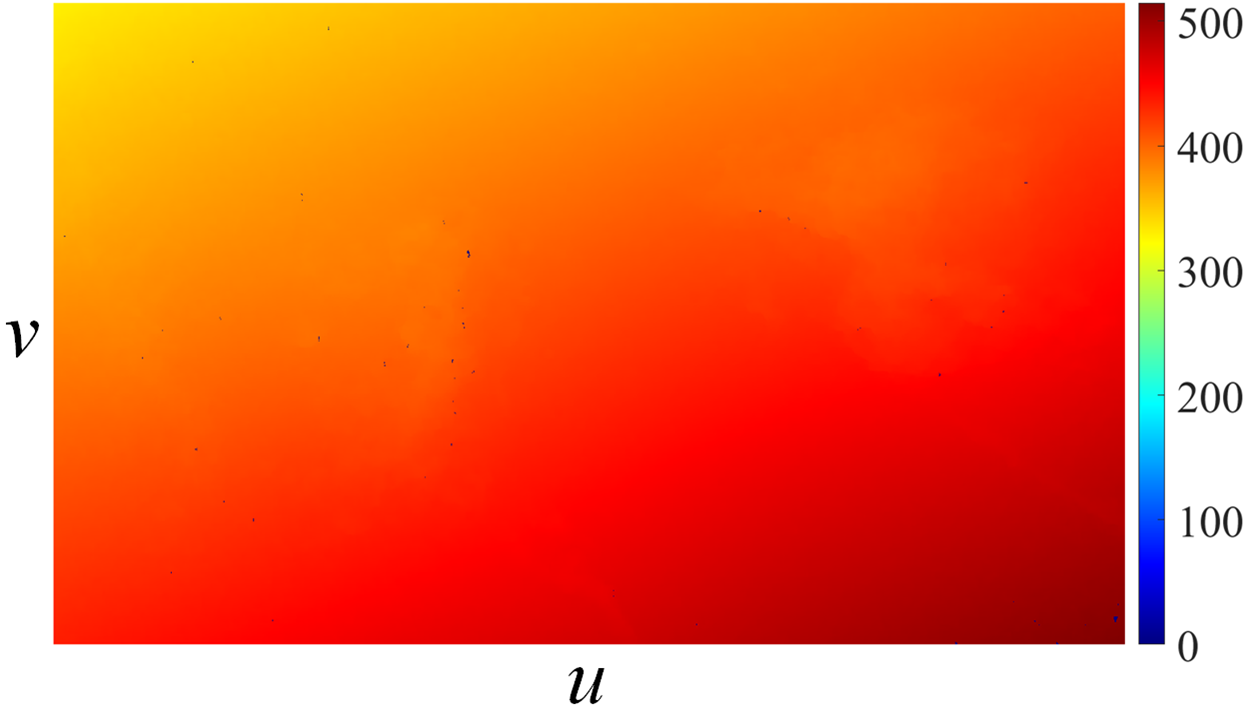}
		\label{fig.disp_map1}
	}
	\subfigure[]{
		\includegraphics[width=0.097\textwidth]{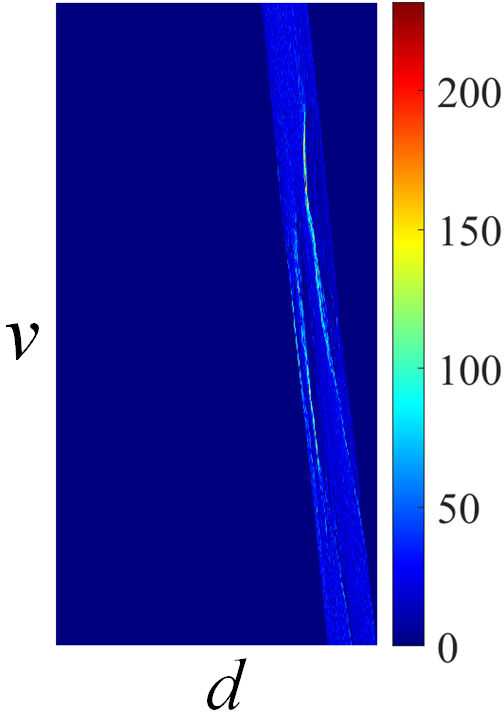}
		\label{fig.v_disp_map1}
	}
	\caption{Disparity map when the roll angle does not equal zero: (a) stereo road images; (b) disparity map; (c) v-disparity map. }
	\label{fig.unwell_roll_angle}
\end{figure}

\section{Pothole Detection Algorithm}
\label{sec.algorithm_description}
The block diagram of the proposed pothole detection algorithm is illustrated in Fig. \ref{fig.block_diagram}, where the algorithm consists of three main components: a) disparity transformation; b) undamaged road area extraction; c) disparity map modeling and pothole detection. 
\subsection{Disparity Transformation}
\label{sec.disp_transformation}
The input of this procedure is a dense disparity map having subpixel accuracy. Since the performance of disparity map modeling relies entirely on the disparity estimation accuracy,  
the dense disparity map was obtained from a stereo road image pair (see Fig. \ref{fig.stereo_img}) through our disparity estimation algorithm \cite{Fan2018}, where the stereo matching search range propagates iteratively from the bottom of the image to the top, and the subpixel disparity map is refined by iteratively minimizing an energy function with respect to the interpolated correlation parabolas. The disparity map is shown in Fig. \ref{fig.disp_map1}, and its corresponding v-disparity map is shown in Fig. \ref{fig.v_disp_map1}. A v-disparity map can be created by computing the histogram of each horizontal row of the disparity map. The proposed pothole detection algorithm is based on the work presented in \cite{Mikhailiuk2016}, where the disparities of the undamaged road surface are modeled by a quadratic surface as follows:
\begin{equation}
g(u,v)=c_0 +c_1 u+ c_2 v+ c_3 u^2+ c_4 v^2+c_5 uv,
\label{eq.quadratic_surface_model}
\end{equation}
where $u$ and $v$ are the horizontal and vertical disparity map coordinates, respectively. The origin of the coordinate system in (\ref{eq.quadratic_surface_model}) is at the center of the disparity map.
Since in our experiments the stereo rig is mounted at a relatively low height, the curvature of the reconstructed road surface is not very high. This makes the values of $c_1$, $c_3$ and $c_5$ in  (\ref{eq.quadratic_surface_model}) very close to zero, when the stereo rig is perfectly parallel to the horizontal road surface. In this case, the projection of the road disparities on the v-disparity map can be assumed to be a parabola of the form \cite{Ozgunalp2017}: 
\begin{equation}
g(v)=\alpha_0+\alpha_1v+\alpha_2v^2.
\label{eq.disparity_projection_model}
\end{equation}

\begin{figure}[!t]
	\begin{center}
		\centering
		\includegraphics[width=0.26\textwidth]{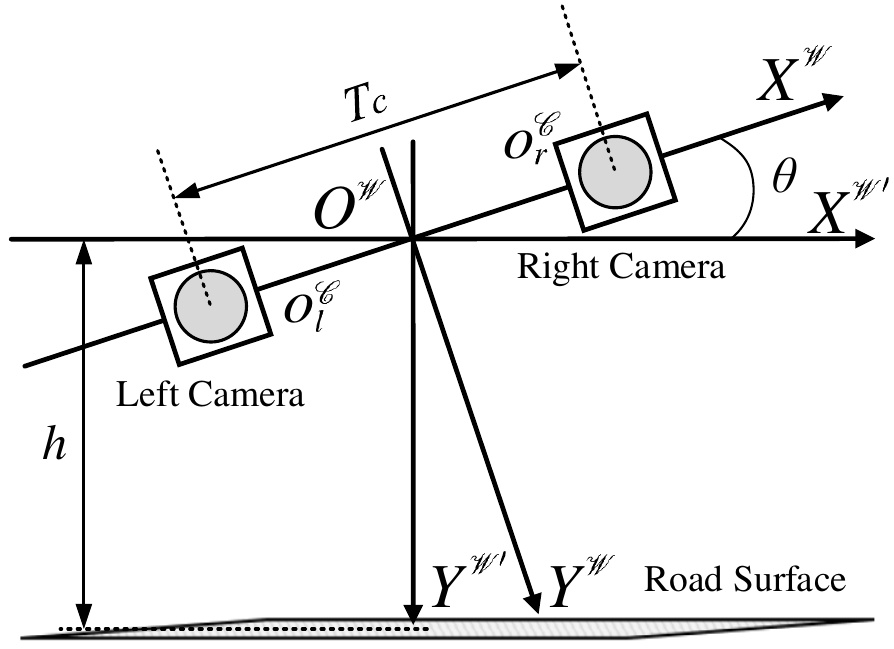}
		\centering
		\caption{Roll angle definition in a stereo vision system.  }
		\label{fig.roll_angle}
	\end{center}
\end{figure}

However, in practice, the stereo rig baseline is not always perfectly parallel to the horizontal road surface. This fact can introduce a non-zero roll angle $\theta$ (see Fig. \ref{fig.roll_angle}) into the imaging process, where $T_c$ and $h$ represent the baseline and the height of the stereo rig, respectively. ${o}_{l}^{\mathscr{C}}$ and ${o}_{r}^{\mathscr{C}}$ are the origins of the left and right camera coordinate systems, respectively. ${O}^{\mathscr{W}}$ is the origin of the world coordinate system. An example of the resulting disparity map is shown in Fig. \ref{fig.disp_map1}, where readers can clearly see that the disparity values change gradually in the horizontal direction, which makes the approach of representing the disparity projection using  (\ref{eq.disparity_projection_model}) somewhat problematic. In this regard, we first estimate the value of the roll angle. The effects caused by the non-zero roll angle are then eliminated by rotating the disparity map by $\theta$.
Finally, the coefficients of the disparity projection model in (\ref{eq.disparity_projection_model}) are estimated, and the disparity map is transformed to better distinguish between damaged and undamaged road areas.

\subsubsection{$\theta$ Estimation and Disparity Map Rotation}
\label{sec.theta_estimation_disparity_map_rotation}
Over the past decade, considerable effort has been made to improve the roll angle estimation. The most commonly used device for this task is an inertial measurement unit (IMU). An IMU can measure the angular rate of a vehicle by analyzing the data acquired using different sensors, such as accelerometers, gyroscopes and magnetometers \cite{EricTseng2007, Oh2013}. In these approaches, the road bank angle is always assumed to be zero, and only the roll angle is considered in the estimation process. However, the estimation of both these two angles is always required in many real-world applications. Unfortunately, this cannot be realized with the use of only IMUs \cite{Ryu2002, Ryu2004}. 

In recent years, many authors  have turned their focus towards estimating the roll angle from disparity maps \cite{Labayrade2003, Skulimowski, Ozgunalp2017, Evans2018, Fan2018c}. For example,   the road surface is assumed to be a horizontal ground plane, and an effective roll angle estimation algorithm was proposed based on v-disparity map analysis \cite{Labayrade2003, Skulimowski}. In \cite{Ozgunalp2017}, the disparities in a selected small area were modeled by a plane $g(u,v)=c_0+c_1u+c_2v$. The roll angle was then calculated as $\arctan(-c_1/c_2)$. However, finding a proper disparity map area for plane fitting is always challenging, because the selected area may contain an obstacle or a pothole, which can severely affect the fitting accuracy \cite{Evans2018}. Furthermore, the above-mentioned algorithms are only suitable for planar road surfaces. Hence, in this subsection, we introduce a roll angle estimation algorithm, which can work effectively for both planar and non-planar road surfaces.

When the roll angle is equal to zero, the vector  $\boldsymbol{\alpha}=[\alpha_0,\alpha_1,\alpha_2]^\top$, storing the disparity projection model coefficients, can be estimated by solving a least squares problem as follows:
\begin{equation}
\begin{split}
\boldsymbol{\alpha}={\argminA_{\boldsymbol{\alpha}}}\ E_0,
\end{split}
\label{eq.estimate_road_model}
\end{equation}
where
\begin{equation}
E_0={\boldsymbol{e}}^\top\boldsymbol{e},
\label{eq.E_formula}
\end{equation}
and
\begin{equation}
\boldsymbol{e}=\boldsymbol{d}-\boldsymbol{V}\boldsymbol{{\alpha}}.
\label{eq.e_formula}
\end{equation}
The column vector
$\boldsymbol{d}=[d_0,\ d_1,\ \cdots,\ d_n]^\top$ stores the disparity values.  $\boldsymbol{V}$ is a matrix of size $(n+1)\times 3$
 given as follows: 
\begin{equation}
\begin{split}
\boldsymbol{V}=
\begin{bmatrix}
1 & v_0 & {v_0}^2\\
1 & v_1 & {v_1}^2\\
\vdots & \vdots & \vdots\\
1 & v_n & {v_n}^2\\
\end{bmatrix}.
\end{split}
\label{eq.V_matrix}
\end{equation}
 This optimization problem has a closed form solution:
\begin{equation}
\boldsymbol{\alpha}=(\boldsymbol{V}^\top\boldsymbol{V})^{-1}\boldsymbol{V}^\top\boldsymbol{d}.
\label{eq.alpha_2}
\end{equation}
The minimum energy $E_{0_\text{min}}$ can also be obtained by combining (\ref{eq.E_formula}), (\ref{eq.e_formula}) and (\ref{eq.alpha_2}):
\begin{equation}
E_{0_\text{min}}=\boldsymbol{d}^\top\boldsymbol{d}-\boldsymbol{d}^\top\boldsymbol{V}(\boldsymbol{V}^\top\boldsymbol{V})^{-1}\boldsymbol{V}^\top\boldsymbol{d}.
\label{eq.E_min}
\end{equation}
However, when the roll angle does not equal zero, the disparity distribution on each row becomes less compact (see Fig. \ref{fig.v_disp_map1}). This greatly affects the accuracy of least squares fitting and produces a much higher $E_{0_\text{min}}$.

\begin{figure}[!t]
	\begin{center}
		\centering
		\includegraphics[width=0.36\textwidth]{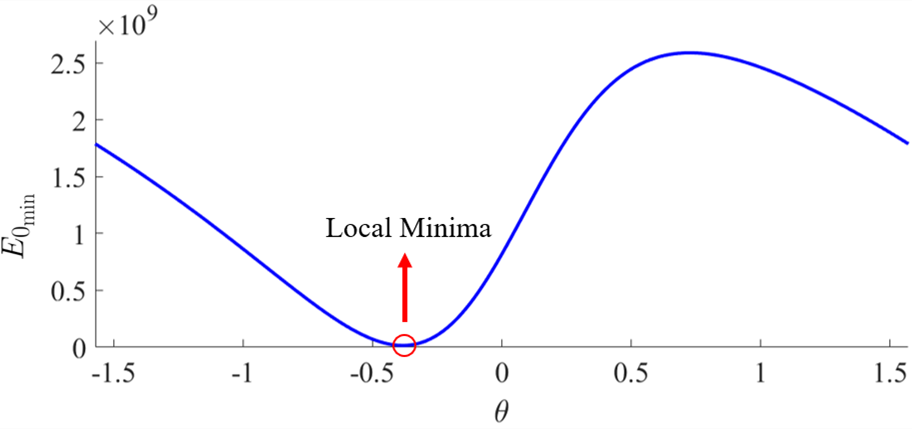}
		\centering
		\caption{$E_{0_\text{min}}$ function versus $\theta$. }
		\label{fig.theta_curve}
	\end{center}
\end{figure}

To rotate the disparity map around a given angle $\theta$, each set of original coordinates $[u,v]^\top$ is transformed to a set of new coordinates $[x,y]^\top$ as follows \cite{Fan2018c}: 
\begin{equation}
\begin{split}
x=u\cos\theta+v\sin\theta,\\
y=v\cos\theta-u\sin\theta.
\end{split}
\label{eq.u_v_transform}
\end{equation}
(\ref{eq.e_formula}) can now be rewritten as follows:
\begin{equation}
\boldsymbol{e}(\theta)=\boldsymbol{d}-\boldsymbol{Y}(\theta)\boldsymbol{{\alpha}}(\theta),
\label{eq.e2_formula}
\end{equation}
where $\boldsymbol{Y}(\theta)$ is an $(n+1)\times 3$ matrix: 
\begin{equation}
\begin{split}
\boldsymbol{Y}(\theta)=
\begin{bmatrix}
1 & y_0(\theta) & {{y_0}(\theta)}^2\\
1 & y_1(\theta) & {{y_1}(\theta)}^2\\
\vdots & \vdots & \vdots\\
1 & y_n(\theta) & {{y_n}(\theta)}^2\\
\end{bmatrix}.
\end{split}
\label{eq.Y_matrix}
\end{equation}
(\ref{eq.alpha_2}) is rewritten as follows:
\begin{equation}
\boldsymbol{\alpha}(\theta)=(\boldsymbol{Y}(\theta)^\top\boldsymbol{Y}(\theta))^{-1}\boldsymbol{Y}(\theta)^\top\boldsymbol{d}.
\label{eq.alpha_3}
\end{equation}
(\ref{eq.E_formula}), (\ref{eq.e2_formula}) and (\ref{eq.alpha_3}) result in the following expression:
\begin{equation}
E_{0_\text{min}}(\theta)=\boldsymbol{d}^\top\boldsymbol{d}-\boldsymbol{d}^\top\boldsymbol{Y}(\theta)(\boldsymbol{Y}(\theta)^\top\boldsymbol{Y}(\theta))^{-1}\boldsymbol{Y}(\theta)^\top\boldsymbol{d}.
\label{eq.E_min2}
\end{equation}
Therefore, the main consideration of the proposed roll angle estimation algorithm is to rotate the disparity map at different angles, and find the angle which minimizes $E_{0_\text{min}}$.
$E_{0_\text{min}}$ with respect to different $\theta$ is illustrated in Fig. \ref{fig.theta_curve}. 
Giving a set of coordinates $[u,v]^\top$, the new coordinate $y$ can be calculated using (\ref{eq.u_v_transform}). The corresponding $E_{0_\text{min}}$ can be computed from  (\ref{eq.E_min2}). Due to the fact that $\cos(\theta+\pi)=-\cos\theta$ and $\sin(\theta+\pi)=-\sin\theta$, the disparity maps rotated around $\theta$ and $\theta+\pi$ are symmetric with respect to the origin of the coordinate system. Namely, (\ref{eq.E_min2}) outputs the same $E_{0_\text{min}}$ no matter how the disparity map is rotated around $\theta$ or $\theta+\pi$. Therefore, we set the interval of $\theta$  to $(-\pi/2,\pi/2]$. The estimation of  $\theta$ is achieved by finding the position of the local minima between $-\pi/2$ and $\pi/2$.

However, finding the local minima is a computationally intensive task, because it involves performing the necessary calculations through the whole interval of $\theta$. Furthermore, the step size $\varepsilon_{\theta}$ has to be set to a very small and practical
value, in order to obtain an accurate value of $\theta$ \cite{Fan2018c}. 
Hence in this paper, GSS is utilized to reduce the searching times. 
The procedure of the proposed $\theta$ estimation algorithm is given in Algorithm \ref{al.roll_angle_estimation}, where $\kappa=(\sqrt{5}-1)/2$ is the golden section ratio \cite{Pedregal2006}.  

\begin{figure}[!t]
	\centering
	\subfigure[]{
		\includegraphics[width=0.262\textwidth]{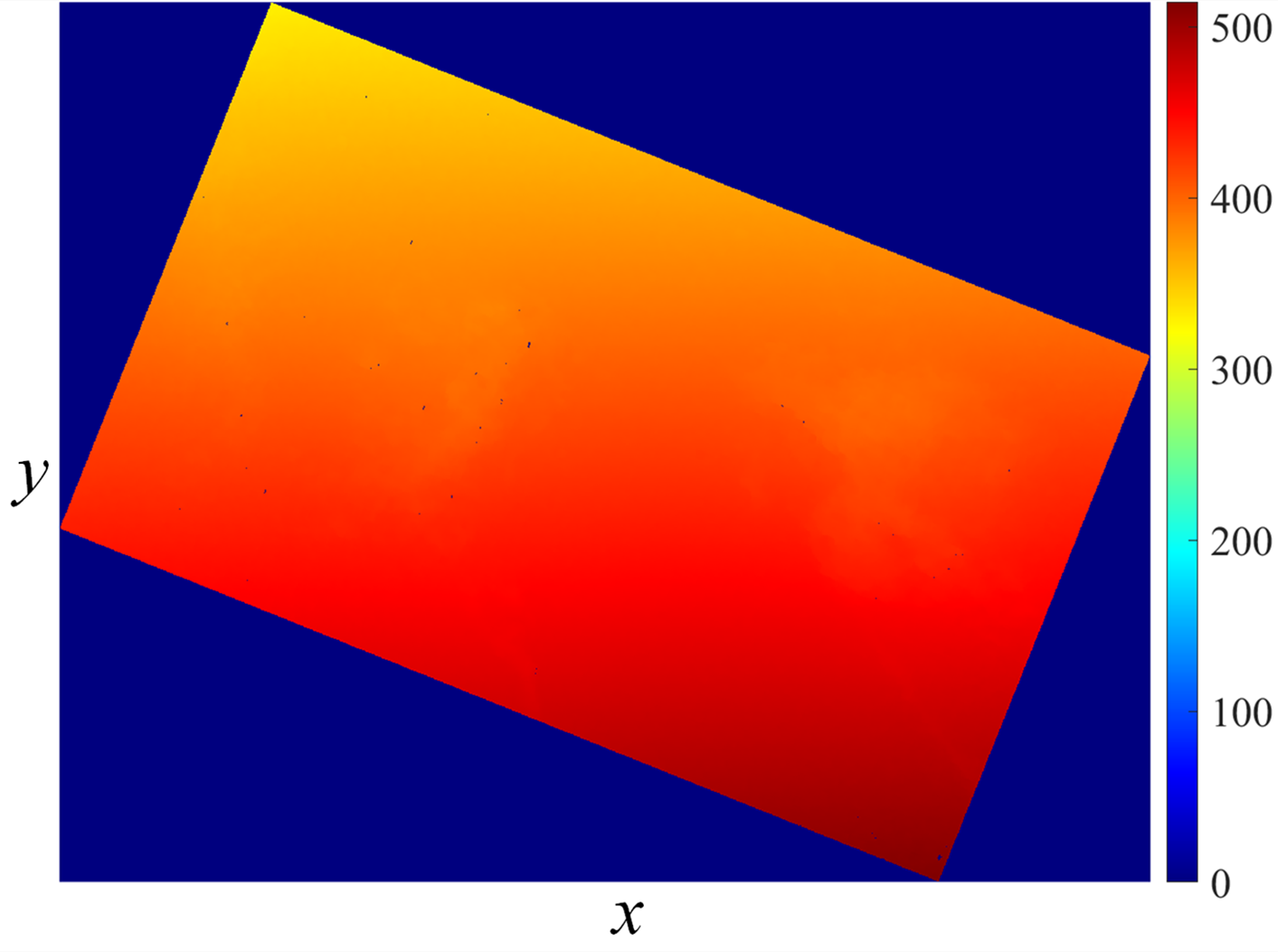}
		\label{fig.disp_map2}
	}
	\subfigure[]{
		\includegraphics[width=0.094\textwidth]{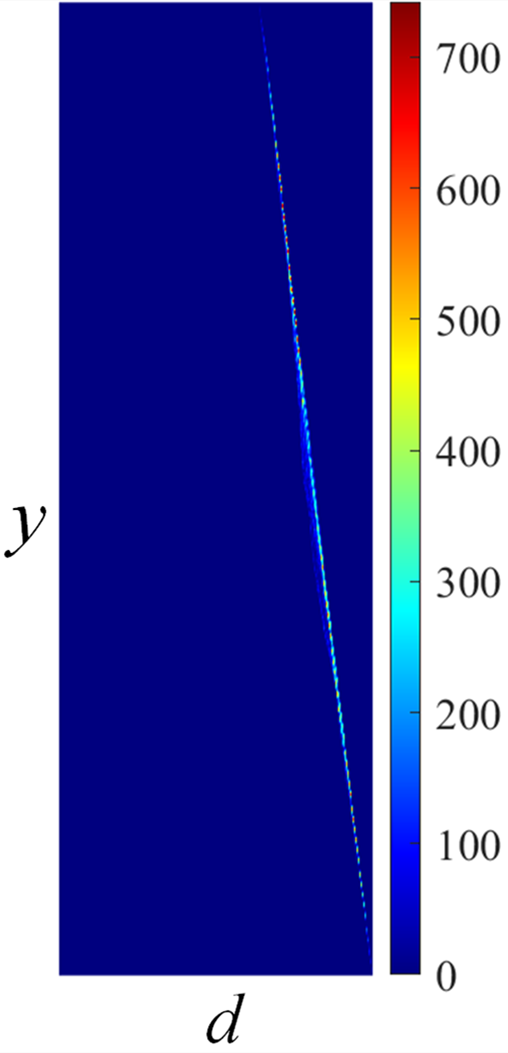}
		\label{fig.v_disp_map2}
	}
	\caption{Elimination of the effects caused by the non-zero roll angle: (a) rotated disparity map; (b) the y-disparity map of Fig. \ref{fig.disp_map2}. }
	\label{fig.well_roll_angle}
\end{figure}

The disparity map is then rotated around $\theta$, as illustrated in Fig. \ref{fig.disp_map2}. A y-disparity map (see Fig. \ref{fig.v_disp_map2}) can be created by computing the histogram of each horizontal row of the rotated disparity map. We can observe that the disparity values on each row become more uniform. The evaluation of the proposed roll angle estimation algorithm will be discussed in Section \ref{sec.roll_angle_evaluation}.

\subsubsection{$\boldsymbol{\alpha}$ Estimation and Disparity Transformation}
\label{sec.alpha_estimation_disparity_transformation}
In this subsection, we utilize DP \cite{Ozgunalp2017} to extract the road disparity projection model  from the y-disparity map. For the purpose of convenient notation, the projection model is also referred to as the target path. The energy of every possible solution is first computed as follows:
\begin{equation}
\begin{split}
E_1(d,y)=-m(d,y)+\min_{\tau}[E_{1}(d+1,y+\tau)+\lambda\tau]\\  \text{s.t.} \  \tau\in[\tau_\text{min},\tau_\text{max}],
\end{split}
\label{eq.DP}
\end{equation}
where $m(d,y)$ represents the y-disparity value at the  position $[d,y]^\top$ and $\lambda$ is a positive smoothness term \cite{Ozgunalp2017}. $E_{1}$ represents the energy of a possible target path in the y-disparity map. $\tau_\text{min}$ is typically set to 0. $\tau_\text{max}$ depends entirely on $\boldsymbol{{\alpha}}$, and it is set to 10 in this paper. 
 The target path $M=\{(d_i,y_i),\ i=0,1,\dots,n\}$ can be found by minimizing the energy function in (\ref{eq.DP}), where $(d_i,y_i)$ stores the horizontal and vertical coordinates of the target path, respectively. 

\begin{algorithm}[t!]
	\KwData{disparity map}
	\KwResult{$\theta$}
	set $\theta_1$ and $\theta_2$ to $-\pi/2$ and $\pi/2$, respectively\;
	compute $E_{0_\text{min}}(\theta_1)$ using Eq. \ref{eq.E_min2}\;
	compute $E_{0_\text{min}}(\theta_2)$ using Eq. \ref{eq.E_min2}\;
	\While{$\theta_2-\theta_1> \varepsilon_{\theta}$}{
		set $\theta_3$ and $\theta_4$ to $\kappa\theta_1+(1-\kappa)\theta_2$ and $\kappa\theta_2+(1-\kappa)\theta_1$, respectively\;
		compute $E_{0_\text{min}}(\theta_3)$ using Eq. \ref{eq.E_min2}\;
		compute $E_{0_\text{min}}(\theta_4)$ using Eq. \ref{eq.E_min2}\;
		\uIf{$E_{0_\text{min}}(\theta_3)>E_{0_\text{min}}(\theta_4)$}{
			$\theta_1$ is replaced by $\theta_3$\;}
		\Else{$\theta_2$ is replaced by $\theta_4$\;}{
		}
	}
	\caption{$\theta$ estimation using GSS.}
	\label{al.roll_angle_estimation}
\end{algorithm}

By substituting the horizontal and vertical coordinates of the target path into  (\ref{eq.estimate_road_model}), (\ref{eq.E_formula}), (\ref{eq.e2_formula}) and (\ref{eq.Y_matrix}), we can obtain $\boldsymbol{{\alpha}}$. The disparity map can, therefore, be transformed using $\theta$ and $\boldsymbol{\alpha}$. However, the outliers in the target path may greatly affect the accuracy of $\boldsymbol{{\alpha}}$ estimation. We, therefore, use RANSAC to update the values in $\boldsymbol{{\alpha}}$. The full list of procedures involved for $\boldsymbol{\alpha}$ estimation are detailed in Algorithm \ref{al.dp_lsf}.
\begin{algorithm}[t!]
	\SetKwInOut{Input}{Input}
	\SetKwInOut{Output}{Output}
	\Input{$M$}
	\Output{$\boldsymbol{{\alpha}}$}
	create a $t\times (s+3)$ matrix $\boldsymbol{T}$\;
	\For{$i$ $\gets$ $1$ \text{to} $t$}{
		randomly select $p$ pairs of coordinates from $M$\;
		estimate $\boldsymbol{\alpha}$ using Eq. \ref{eq.estimate_road_model}\; 
		$\boldsymbol{T}(i,s+1:s+3)\gets\boldsymbol{{\alpha}}^\top$\;
		\For{$j\gets1$ to $s$}{
			set the tolerance to  $\varepsilon_{\boldsymbol{{\alpha}}}/2^{j-1}$\;
			compute the number of inliers $n_\text{inlier}$\;
			compute the number of outliers $n_\text{outlier}$\;
			compute the ratio  $\eta=n_{\text{inlier}}/n_\text{outlier}$\;
			$\boldsymbol{T}(i,j)\gets{{\eta}}$\;
		}{
		}
	}
	$j=0$\;
	\Do{the highest $\eta$ in  the $j$th column of $\boldsymbol{T}$ corresponds to  more than one $\boldsymbol{{\alpha}}$}
	{
		$j\gets j+1$\;
		find the highest $\eta$ in the $j$th column of $\boldsymbol{T}$\;	
	}
	find $\boldsymbol{{\alpha}}$ which corresponds to  the highest $\eta$ in the $j$th column of $\boldsymbol{T}$\;	
	\caption{$\boldsymbol{\alpha}$ estimation using DP.}
	\label{al.dp_lsf}
\end{algorithm}

RANSAC is iterated $t$ times. Selecting a higher $t$ raises the possibility of finding the best $\boldsymbol{\alpha}$ but also increases the processing time. In order to minimize the trade-off between speed and robustness, $t$ is set to $50$ in this paper. In each iteration, we select $p$ pairs of coordinates $[d,y]^\top$ from the target path to estimate $\boldsymbol{{\alpha}}$. For a smaller $p$, there is less chance that any outliers will influence optimization. In this paper $p$ is set to $3$, which is the smallest possible value for determining $\boldsymbol{{\alpha}}$. The ratio $\eta$ of inliers versus outliers can then be computed with respect to a given tolerance $\varepsilon_{\boldsymbol{{\alpha}}}$. 
The best $\boldsymbol{{\alpha}}$ corresponds to the highest ratio $\eta$. However, the selection of an appropriate  $\varepsilon_{\boldsymbol{{\alpha}}}$ can be challenging, as it is possible that there could be more than one satisfying value for $\boldsymbol{{\alpha}}$. Hence, in this paper, the value of $\varepsilon_{\boldsymbol{{\alpha}}}$ also changes $s$ times. In each iteration, the value of $\varepsilon_{\boldsymbol{{\alpha}}}$ reduces by half and $\eta$ is computed. The best $\boldsymbol{{\alpha}}$ can be determined by finding the highest ratio $\eta$. In this paper, the values of $\varepsilon_{\boldsymbol{{\alpha}}}$ and $s$ are both set to $4$. 

Finally, each disparity is transformed using the following equation:
\begin{equation}
\tilde{d}(u,v,d,\theta)=[1\ y(u,v,\theta)\ {y(u,v,\theta)}^2]\boldsymbol{{\alpha}}(\theta)-d+\delta,
\label{eq.disp_transformation}
\end{equation}
where $\delta$ is a constant value set to guarantee that all the transformed disparity values are non-negative. In this paper, we set $\delta$ to $30$. The transformed disparity map is shown in Fig. \ref{fig.trans_disp_map}. It can be clearly seen that the disparity values in the undamaged road areas become more uniform, while they differ significantly from those in the damaged areas (potholes). This makes the extraction of undamaged road regions much simpler. 

\subsection{Undamaged Road Area Extraction}
\label{sec.undamaged_road_area_extraction}
\begin{figure}[!t]
	\centering
	\subfigure[]{
		\includegraphics[width=0.242\textwidth]{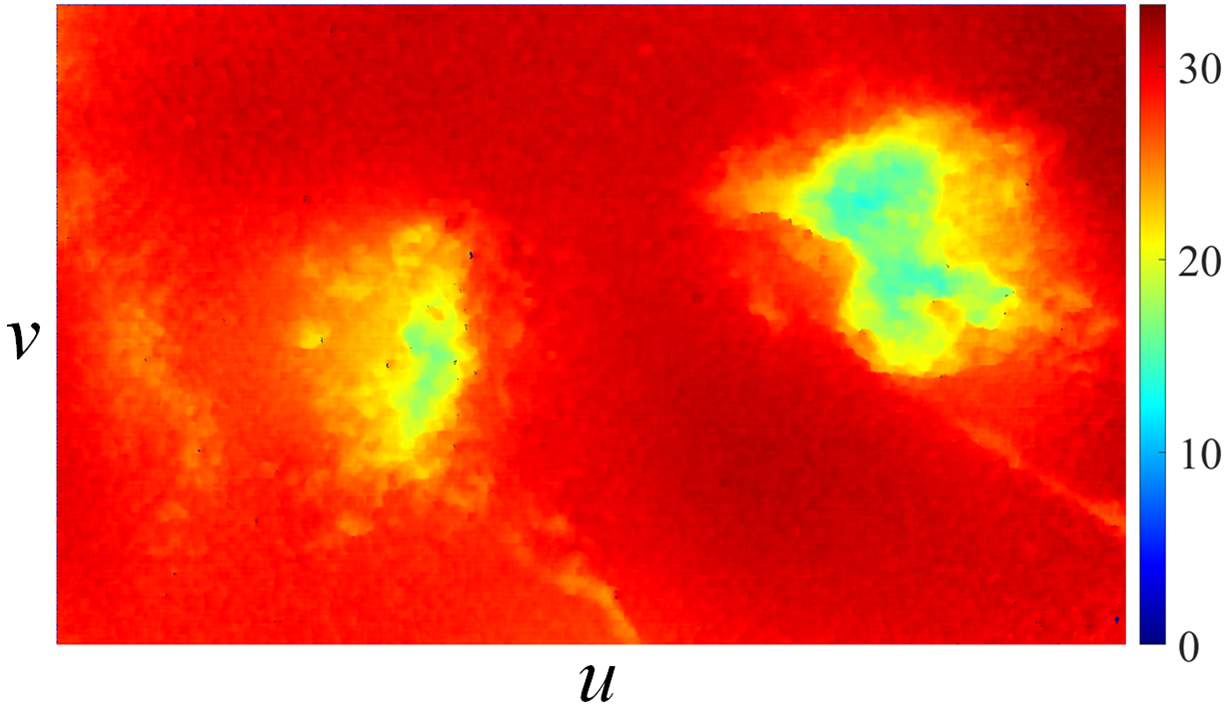}
		\label{fig.trans_disp_map}
	}
	\subfigure[]{
		\includegraphics[width=0.213\textwidth]{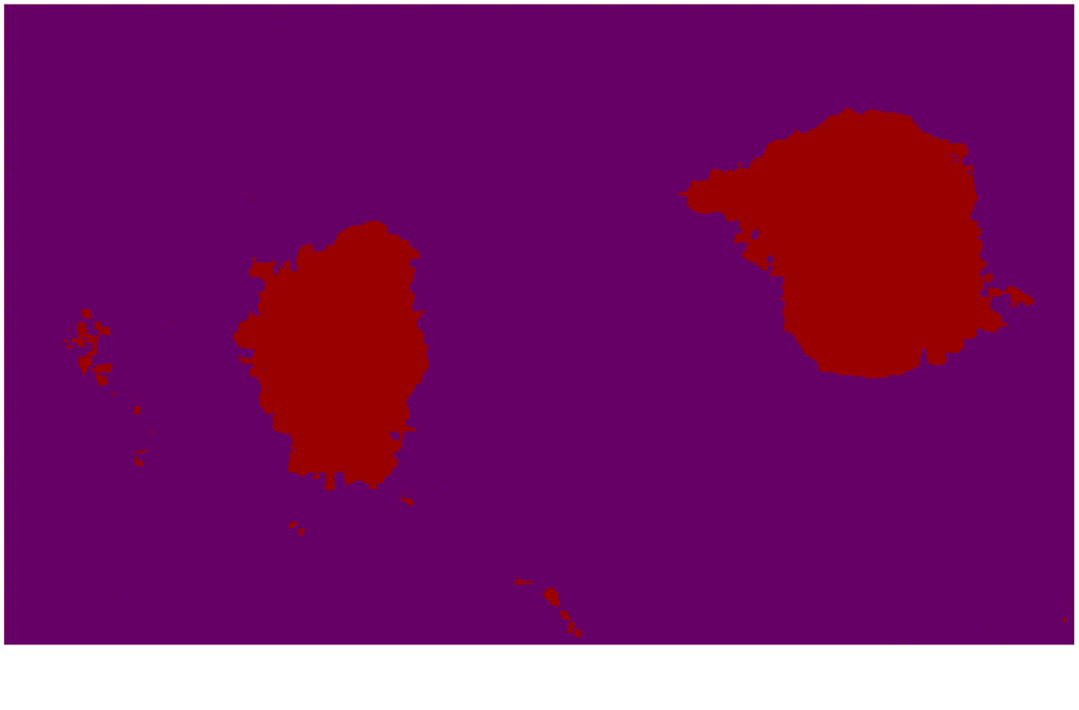}
		\label{fig.segmentation}
	}
	\caption{Disparity transformation and undamaged road area extraction: (a) transformed disparity map; (b) extracted  undamaged road areas. }
	\label{fig.trans_seg}
\end{figure}

Next, we utilize Otsu's thresholding method to segment the transformed disparity map. The segmentation threshold $T_\text{o}$  can be obtained by maximizing the inter-class variance ${\sigma}_\text{o}^2$ as follows \cite{Otsu1979}:
\begin{equation}
{\sigma}_\text{o}^2(T_\text{o})=P_0(T_\text{o})P_1(T_\text{o})[\mu_0(T_\text{o})-\mu_1(T_\text{o})]^2,
\label{eq.otsu}
\end{equation}
where
\begin{equation}
P_0(T_\text{o})=\sum_{i=\tilde{d}_\text{min}}^{T_\text{o}-1}p(i),\ \ P_1(T_\text{o})=\sum_{i=T_\text{o}}^{\tilde{d}_\text{max}}p(i)
\end{equation}
represent the probabilities of damaged and undamaged road areas, respectively. $p(i)$ is the probability of $\tilde{d}=i$. The average disparity values of the damaged and undamaged road areas are given by: 
\begin{equation}
\begin{split}
\mu_0(T_\text{o})=\frac{1}{P_0(T_\text{o})}\sum_{i=\tilde{d}_\text{min}}^{T_\text{o}-1}i p(i),\\
\mu_1(T_\text{o})=\frac{1}{P_1(T_\text{o})}\sum_{i=T_\text{o}}^{\tilde{d}_\text{max}}i p(i).\end{split}
\end{equation}
 The segmentation result is shown in Fig. \ref{fig.segmentation}. We can see that the undamaged road area is successfully extracted. However,  Otsu's thresholding method will always classify the disparities into two categories, even if the transformed disparity map does not contain any road damage. We, therefore, carry out disparity map modeling to ensure that the potholes are correctly detected. 

\subsection{Disparity Map modeling and Pothole Detection}
\label{sec.pothole_detection_labelling}
A common practice in 3D modeling-based pothole detection algorithms \cite{Zhang2014, Mikhailiuk2016} is to fit a quadratic surface to either a 3D point cloud or a 2D disparity map. In parallel axis stereo vision, the point cloud is generated from the disparity map as follows \cite{Ozgunalp2017}:
\begin{equation}
X^{\mathscr{W}}=\frac{u T_c}{d},\ \ Y^{\mathscr{W}}=\frac{v T_c}{d}, \ \ Z^{\mathscr{W}}=\frac{f T_c}{d},\label{eq.duv_xyz}
\end{equation}
where $f$ is the camera focal length.  A disparity error larger than one pixel may result in a non-negligible difference in the point cloud \cite{Haller2012}. Therefore,  disparity map modeling can avoid such errors generated from  (\ref{eq.duv_xyz}), producing greater accuracy compared to point cloud modeling.

To model the disparity map, $\boldsymbol{c}=[c_0,c_1,c_2,c_3,c_4,c_5]^\top$, storing the quadratic surface model  coefficients  can be estimated as follows:
\begin{equation}
\boldsymbol{c}=(\boldsymbol{W}^\top\boldsymbol{W})^{-1}\boldsymbol{W}^\top\boldsymbol{d},
\label{eq.c1}
\end{equation}
where 
\begin{equation}
\begin{split}
\boldsymbol{W}=
\begin{bmatrix}
1 & u_0 & v_0 & {u_0}^2 & {v_0}^2 & {u_0}{v_0}\\
1 & u_1 & v_1 & {u_1}^2 & {v_1}^2 & {u_1}{v_1}\\
\vdots & \vdots & \vdots & \vdots & \vdots & \vdots\\
1 & u_n & v_n & {u_n}^2 & {v_n}^2 & {u_n}{v_n}\\
\end{bmatrix}.
\end{split}
\label{eq.W_matrix}
\end{equation}

However, potential outliers can severely affect the accuracy of disparity map modeling and therefore need to be  discarded beforehand. In this subsection, a disparity map point is determined as an outlier if it fulfills one of the following conditions:
\begin{itemize}
\item it is located in one of the damaged road areas.
\item its surface normal vector differs greatly from the optimal one. 
\item its disparity value differs greatly from the one computed using Eq. \ref{eq.quadratic_surface_model}.
\end{itemize}

In Section \ref{sec.undamaged_road_area_extraction}, the undamaged road areas are successfully extracted and we only use the disparities in this area to model the disparity map. The rest of this subsection presents the approaches for determining the outliers which satisfy the last two conditions and the process of modeling the disparity map without using these outliers. 
\subsubsection{Optimal Normal Vector Estimation}
\label{sec.optimal_normal_vector_estimation}
\begin{figure}[!t]
	\begin{center}
		\centering
		\includegraphics[width=0.38\textwidth]{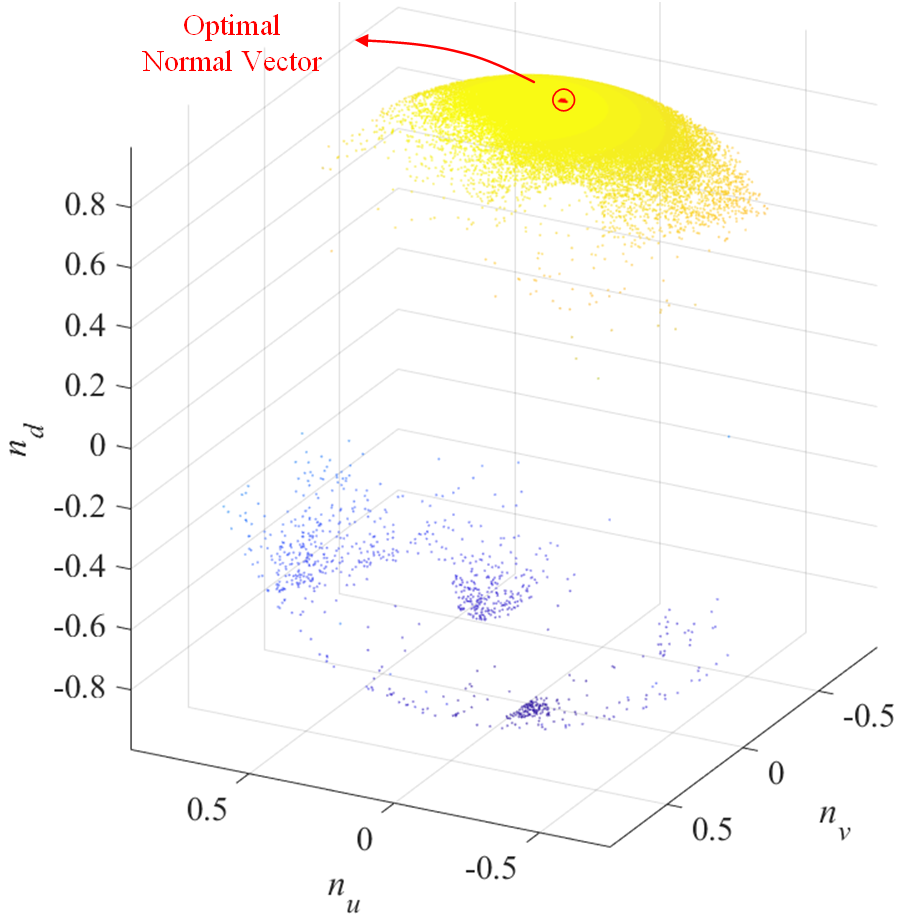}
		\centering
		\caption{Surface normal vectors mapping on a sphere.}
		\label{fig.optimal_normal_vector}
	\end{center}
\end{figure}
For each point $\boldsymbol{p}_i=[u_i,\ v_i,\ d_i]^\top$ in the undamaged road area, we would like to estimate a normal vector $\boldsymbol{n}_i=[n_{ui},\ n_{vi},\ n_{di}]^\top$ from a set of $k$ points in its neighborhood $\boldsymbol{Q}_i=[\boldsymbol{q}_{i1},\  \boldsymbol{q}_{i2},\  \cdots,\  \boldsymbol{q}_{ik}]^\top$, where $\boldsymbol{q}_{ij}\neq\boldsymbol{p}_i$. Here, we define the augmented neighbor matrix  $\boldsymbol{Q}_i^+$ which contains all neighbors and the point $\boldsymbol{p}_i$ itself as follows:
\begin{equation}
\boldsymbol{Q}_i^+=[\boldsymbol{p}_i,\ {\boldsymbol{Q}_i}^\top]^\top.
\label{eq.Q+}
\end{equation}

Existing normal vector estimation methods are generally classified into one of two categories: optimization-based and averaging-based \cite{Klasing2009}.  Although the performance of normal vector estimation  depends primarily  on the application itself,  \textit{PlanePCA} \cite{Wang}, an optimization-based normal vector estimation method, has superior performance in terms of both speed and accuracy. Hence in this subsection, we utilize \textit{PlanePCA} to estimate the normal vectors of the disparities. $\boldsymbol{n}_i$ can be estimated as follows:
\begin{equation}
\boldsymbol{n}_i=\argminA_{\boldsymbol{n}_i}\Big|\Big|
\Big[\boldsymbol{Q}_i^+-\bar{\boldsymbol{Q}}_i^+
\Big]\boldsymbol{n}_i
\Big|\Big|_2,
\label{eq.normal_estimation}
\end{equation}
where 
\begin{equation}
\bar{\boldsymbol{Q}}_i^+=\boldsymbol{1}_{k+1}
{\big(\frac{1}{k+1}{\boldsymbol{Q}_i^+}^\top\boldsymbol{1}_{k+1}\big)}^{\top}.
\label{eq.Q_bar_+}
\end{equation}
$\boldsymbol{1}_{m}$ represents an $m\times1$ vector of ones. Due to the fact that the normal vectors are normalized, they can be projected on a sphere, as shown in Fig. \ref{fig.optimal_normal_vector}, where we can clearly see that the projections are distributed  in a small area.  Therefore, the optimal normal vector $\hat{\boldsymbol{n}}$ can be determined by finding the position at which the projections distribute most intensively. 

Since the projection of $\hat{\boldsymbol{n}}$ is also on the sphere, it can be written in spherical coordinates as follows: 
\begin{equation}
\begin{split}
	\hat{\boldsymbol{n}}=[\sin\varphi_1\cos\varphi_2,\ \sin\varphi_1\sin\varphi_2,\  \cos\varphi_1]^\top\
	\\ \text{s.t.}\ \
\varphi_1\in[0,\pi], \ \ \varphi_2\in[0,2\pi). 
\end{split}
\label{eq.n_sphere}
\end{equation}
It can be estimated by minimizing $E_2$:
\begin{equation}
E_2={\boldsymbol{1}_{n+1}}^\top \boldsymbol{m},
\label{eq.n_hat_estimation}
\end{equation}
where
\begin{figure}[!t]
	\centering
	\subfigure[]{
		\includegraphics[width=0.246\textwidth]{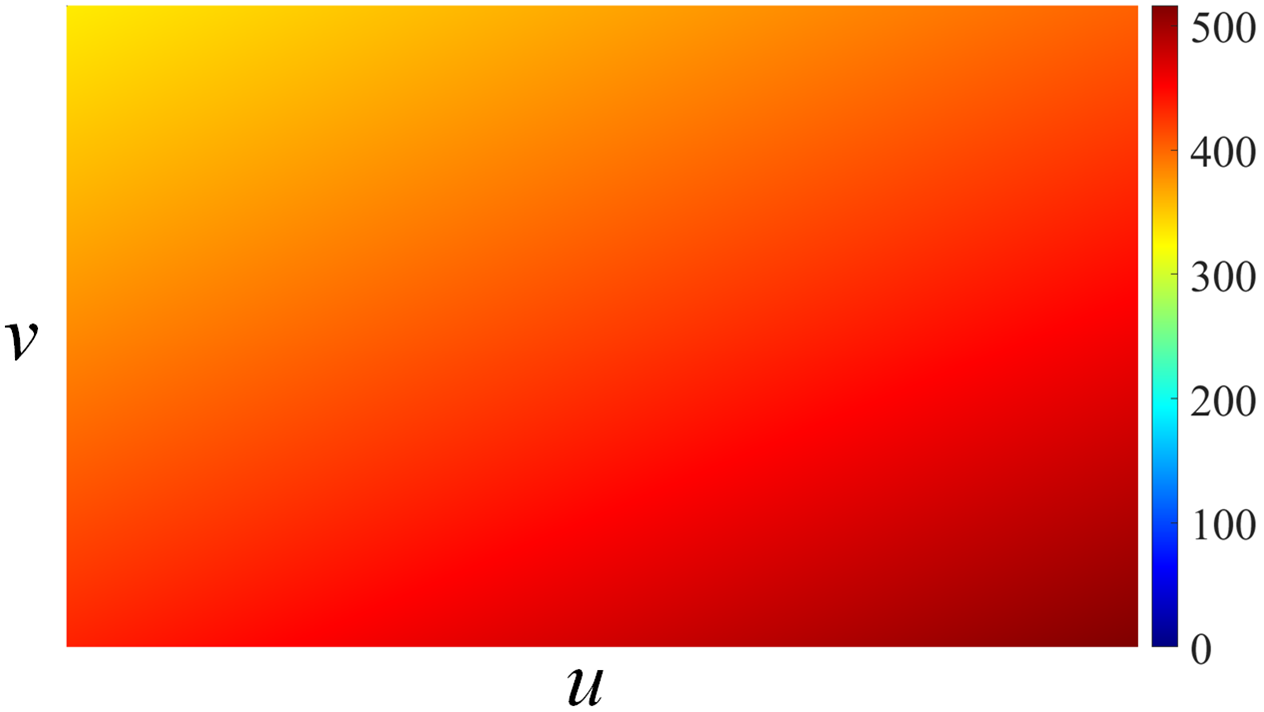}
		\label{fig.modeled_disp_map}
	}
	\subfigure[]{
		\includegraphics[width=0.209\textwidth]{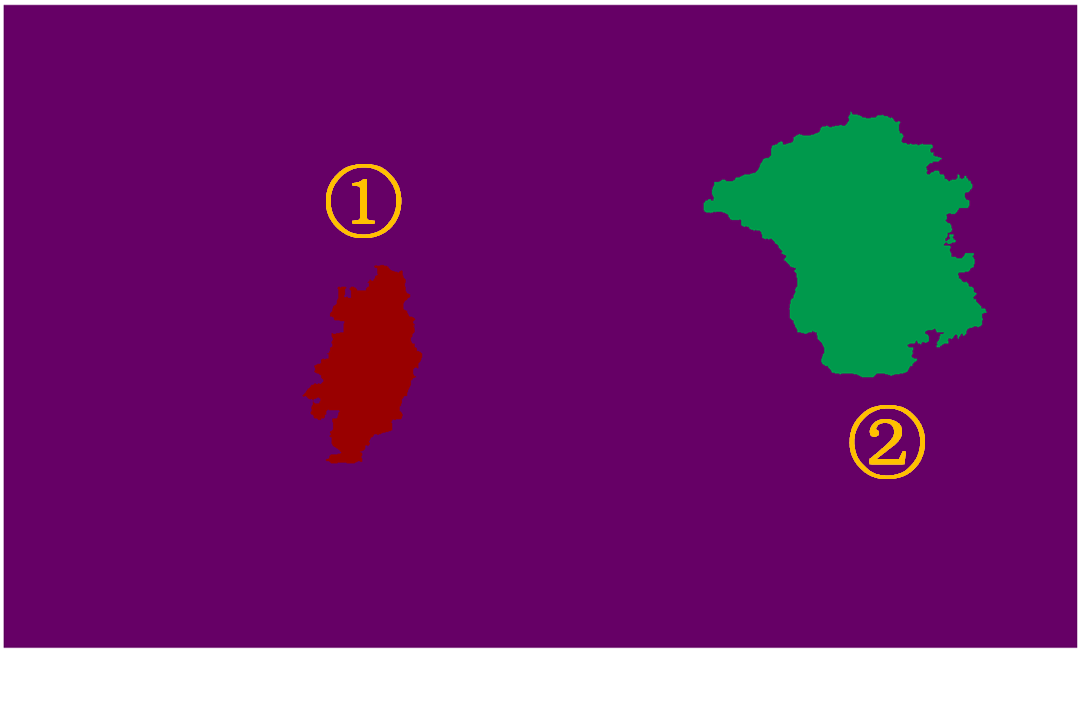}
		\label{fig.detection_result}
	}
	\caption{Disparity map modeling and pothole detection: (a) modeled disparity map, (b) detected potholes.  }
	\label{fig.modeled_disp_detection_result}
\end{figure}
\begin{equation}
\boldsymbol{m}= [-\boldsymbol{n}_{0}\cdot\hat{\boldsymbol{n}},\ -\boldsymbol{n}_{1}\cdot\hat{\boldsymbol{n}},\ \cdots,\ -\boldsymbol{n}_{n}\cdot\hat{\boldsymbol{n}}]^\top.
\label{eq.m}
\end{equation}
By applying (\ref{eq.n_sphere}) and (\ref{eq.m}) to (\ref{eq.n_hat_estimation}), the following expressions are derived:
\begin{equation}
	\tan\varphi_1=\frac{{\boldsymbol{1}_{n+1}}^\top\boldsymbol{n}_{u}\cos\varphi_2+{\boldsymbol{1}_{n+1}}^\top\boldsymbol{n}_{v}\sin\varphi_2}{{\boldsymbol{1}_{n+1}}^\top\boldsymbol{n}_{d}},
	\label{eq.varphi_1}
\end{equation}
\begin{equation}
\tan\varphi_2=\frac{{\boldsymbol{1}_{n+1}}^\top\boldsymbol{n}_{v}}{{\boldsymbol{1}_{n+1}}^\top\boldsymbol{n}_{u}},
\label{eq.varphi_2}
\end{equation}
where $\boldsymbol{n}_{u}=[n_{u0},n_{u1},\cdots,n_{un}]^\top$, $\boldsymbol{n}_{v}=[n_{v0},n_{v1},\cdots,n_{vn}]^\top$ and $\boldsymbol{n}_{d}=[n_{d0},n_{d1},\cdots,n_{dn}]^\top$.
Due to the fact that $\varphi_2$ is between $0$ and $2\pi$,  (\ref{eq.varphi_2}) will have two solutions: 
\begin{equation}
\varphi_2=\arctan\frac{{\boldsymbol{1}_{n+1}}^\top\boldsymbol{n}_{v}}{{\boldsymbol{1}_{n+1}}^\top\boldsymbol{n}_{u}}+k\pi, \ k\in\{0,1\}.
\end{equation}
Substituting each $\varphi_2$ into (\ref{eq.varphi_1})  produces a value for $\varphi_1$. The two pairs of $[\varphi_1,\  \varphi_2]^\top$ correspond to the maxima and  minima of $E_2$, respectively. By substituting each pair of $[\varphi_1,\  \varphi_2]^\top$ into Eq. \ref{eq.n_hat_estimation} and comparing the two obtained values, we can find the optimal normal vector. If the angle between $\hat{\boldsymbol{n}}$ and $\boldsymbol{n}_i$ exceeds a pre-set threshold $\varepsilon_{\boldsymbol{n}}$, the corresponding disparity will be considered as an outlier and will not be used for disparity map modeling. In our experiments, we assume that the second category of outliers account for $10\%$ of the undamaged road areas, and therefore, $\varepsilon_{\boldsymbol{n}}$ is set to $\pi/36$ rad. The outliers satisfying the first two conditions can then be successfully removed. The third category of outliers are removed along with the disparity map modeling.
\subsubsection{Disparity Map Modeling}
\label{sec.disp_map_modeling}
To model the disparity map with more robustness, we use RANSAC to reduce the effects caused by the third category of outliers described in Section \ref{sec.pothole_detection_labelling}. Here, RANSAC is iterated $t$ times. In each iteration, a subset of disparities are  selected randomly to estimate $\boldsymbol{c}$. To ensure uniform distribution of the selected disparities, we equally divide the disparity map into a group of square blocks and select one disparity from each block. The disparity block size is $r\times r$. As $r$ becomes smaller, more disparities will be used for surface fitting, which increases the computational complexity. In contrast, selecting a higher value for $r$ results in less computational complexity, but potentially increases noise sensitivity. In this paper, the value of $r$ is set to $125$, which produces approximately $100$ square blocks for our disparity maps. In each iteration, the differences between the actual and fitted disparities are computed and the ratio $\eta$ between the inliers and outliers are obtained. $\boldsymbol{c}$ which corresponds to the highest $\eta$ is then selected as the desirable surface coefficients. Algorithm \ref{al.dp_lsf} presents more details on the least squares fitting using RANSAC. The modeled disparity map is shown in Fig. \ref{fig.modeled_disp_map}. 
\subsubsection{Pothole Detection}
\label{sec.pothole_detection}
\begin{figure}[!t]
	\centering
	\includegraphics[width=0.32\textwidth]{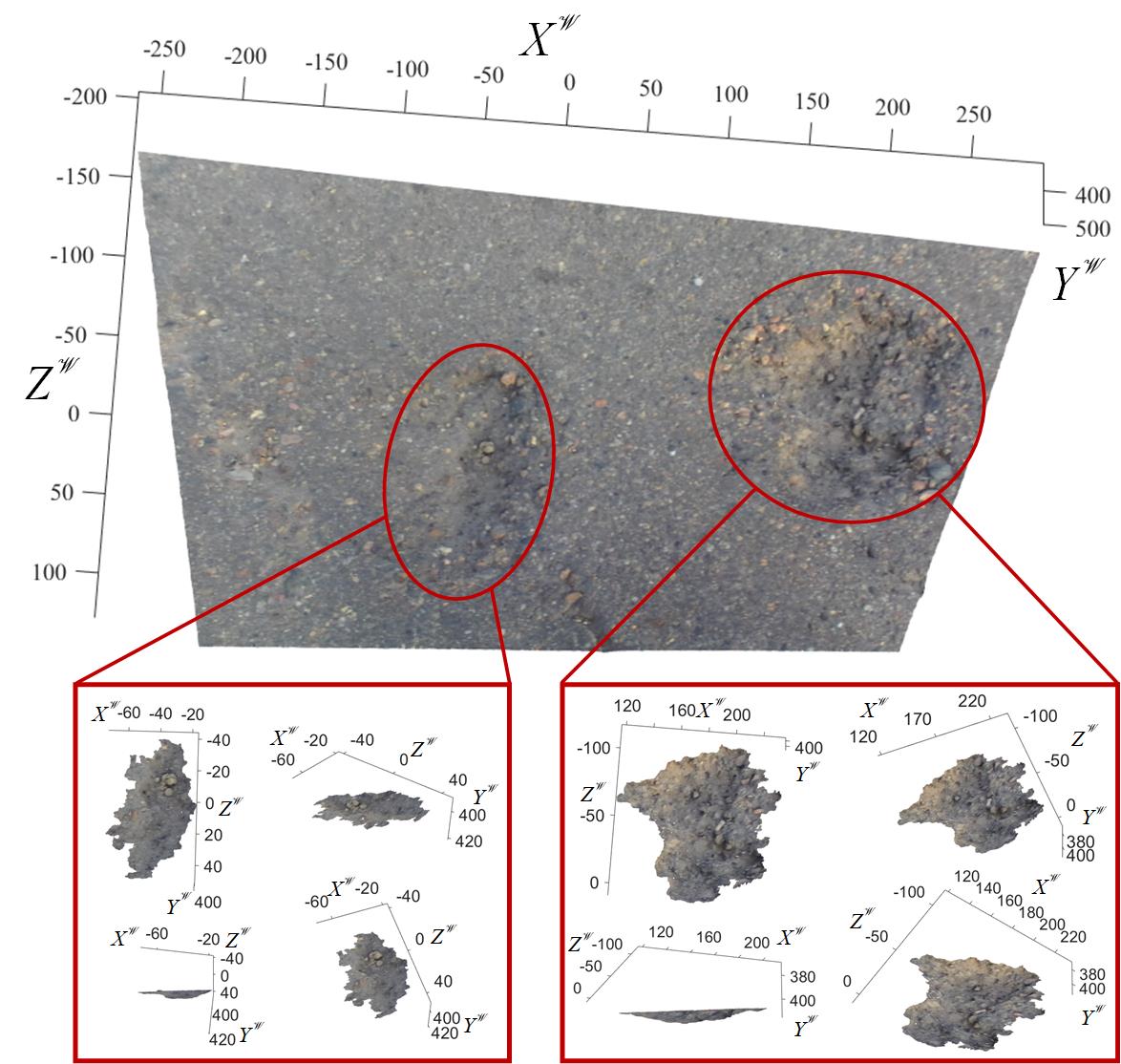}
	\centering
	\caption{The point clouds of the detected potholes.   }
	\label{fig.3d_pc}
\end{figure}
The potholes can then be  detected by finding the regions where the differences between the actual and modeled disparities are larger than a pre-set threshold $\varepsilon_d$. Before labeling different potholes using CCL, the connected components containing fewer than $w$  pixels are removed, because they severely affect the pothole labeling accuracy. Furthermore, the small holes in each connected component are filled, as they are considered to be noise. The selection of $\varepsilon_d$ and $w$ is discussed in Section \ref{sec.pothole_detection_evaluation}. The detected potholes are shown in Fig. \ref{fig.detection_result}.  Finally, the point clouds of the detected potholes are extracted from the reconstructed 3D road surface. The corresponding results are shown in Fig. \ref{fig.3d_pc}.

\section{Experimental Results}
\label{sec.experimental_results}
In this section, the performance of the proposed pothole detection algorithm is evaluated both qualitatively and quantitatively. The proposed algorithm was implemented in MATLAB on an Intel Core i7-8700K CPU using a single thread. The following subsections provide details on the experimental set-up and the evaluation of the proposed algorithm.  
\subsection{Experimental Set-Up}
\label{sec.exp_set_up}
\begin{figure}[!t]
	\centering
	\includegraphics[width=0.22\textwidth]{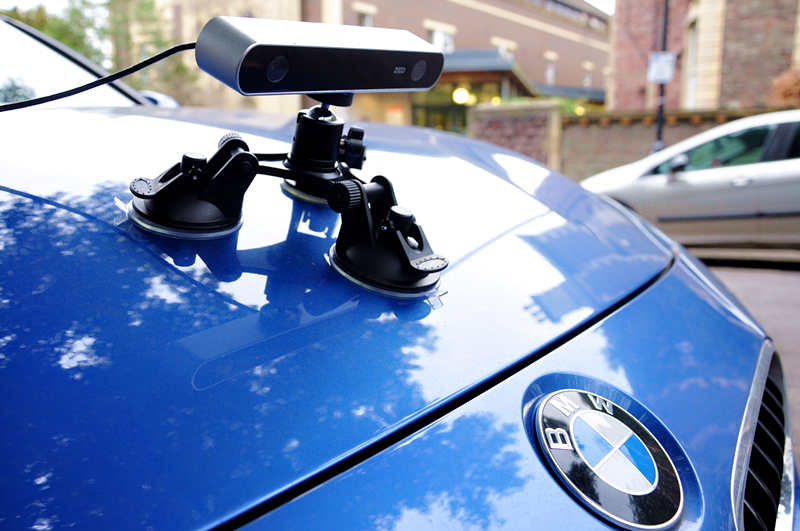}
		\centering
	\caption{Experimental set-up for acquiring stereo road images.   }
	\label{fig.exp_setup}
\end{figure}
In this work, we utilized a ZED stereo camera\footnote{https://www.stereolabs.com/} to capture stereo road images. An example of the experimental set-up is shown in Fig. \ref{fig.exp_setup}. The stereo camera is calibrated manually using the stereo calibration toolbox in MATLAB R2018b. Using the above-mentioned experimental set-up, we created three datasets containing 67 pairs of stereo images. The image resolutions of dataset 1, 2 and 3 are $1028\times1730$, $1030\times1720$, $1028\times1710$ pixels, respectively. The disparity maps are estimated using our previously published algorithm \cite{Fan2018}. All datasets are publicly available and can be found at: \url{ruirangerfan.com}. 

The following subsections analyze the accuracy of roll angle estimation, disparity transformation and pothole detection. 
\subsection{Evaluation of Roll Angle Estimation}
\label{sec.roll_angle_evaluation}

In this subsection, we first analyze the computational complexity of the proposed roll angle estimation algorithm. When estimating the roll angle without using GSS, we have to search through the whole interval of $(-\frac{\pi}{2},\frac{\pi}{2}]$ to find the local minima. Therefore, the computational complexity is $\mathcal{O}(\frac{\pi}{\varepsilon_{\theta}})$. In our method, GSS reduces the interval size exponentially. As a result, the interval size then becomes $\kappa^n\pi$ after the $n$-th iteration. Therefore, the proposed roll angle estimation algorithm reduces the computational complexity to $\mathcal{O}(\log_\kappa \frac{\varepsilon_{\theta}}{\pi})$. 
The proposed roll angle estimation algorithm needs 21  iterations to produce a roll angle, with an accuracy higher than $\frac{\pi}{18000}$ rad.


\begin{figure}[!t]
	\centering
	\includegraphics[width=0.48\textwidth]{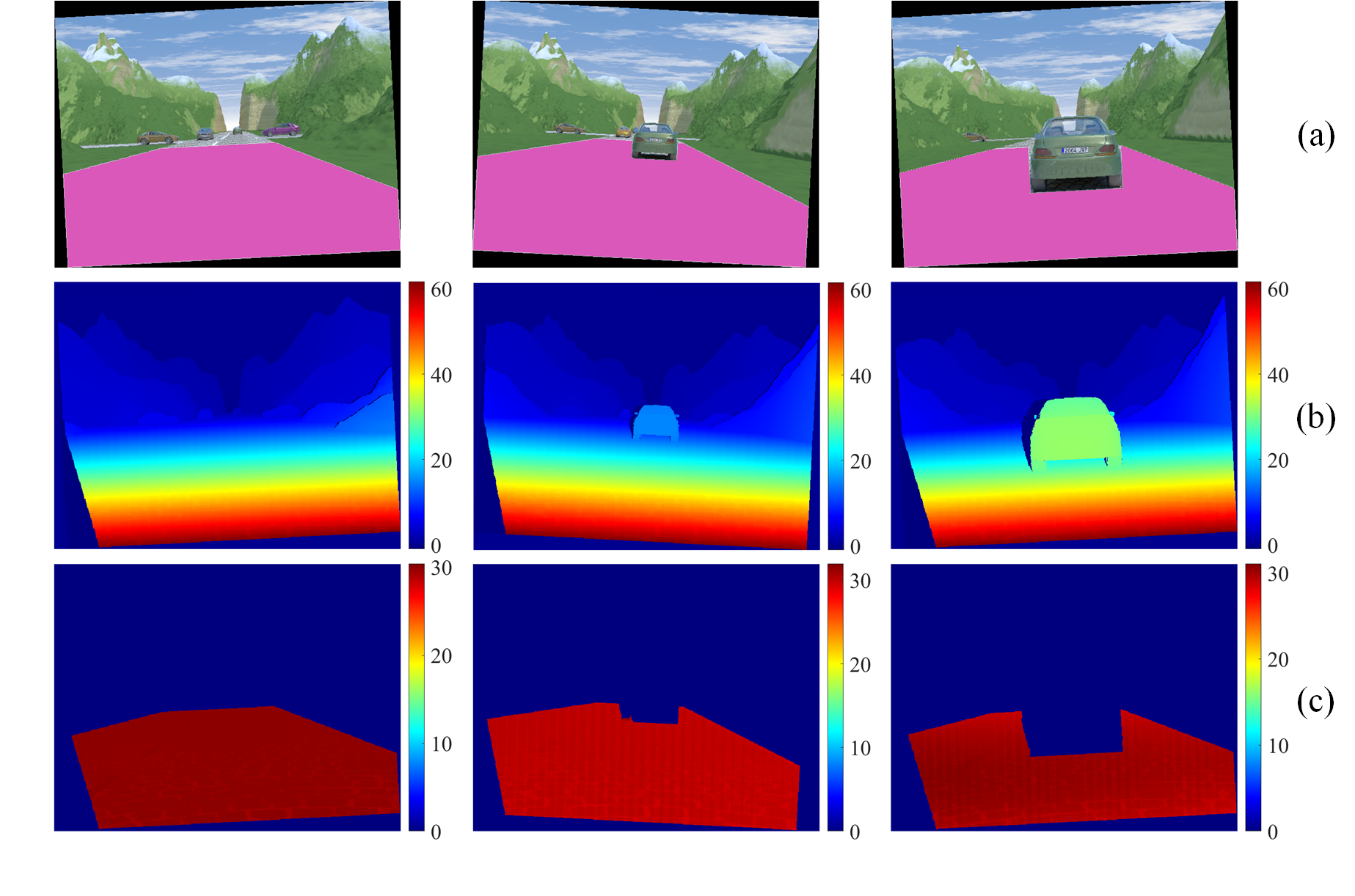}
	\centering
	\caption{Experimental results of the EISATS synthesized stereo dataset: (a) left stereo images (the areas in magenta are our manually selected road areas); (b) ground truth disparity maps; (c) transformed disparity maps.  }
	\label{fig.roll_angle_evaluation}
\end{figure}
\begin{figure*}[!t]
	\centering
	\includegraphics[width=0.9\textwidth]{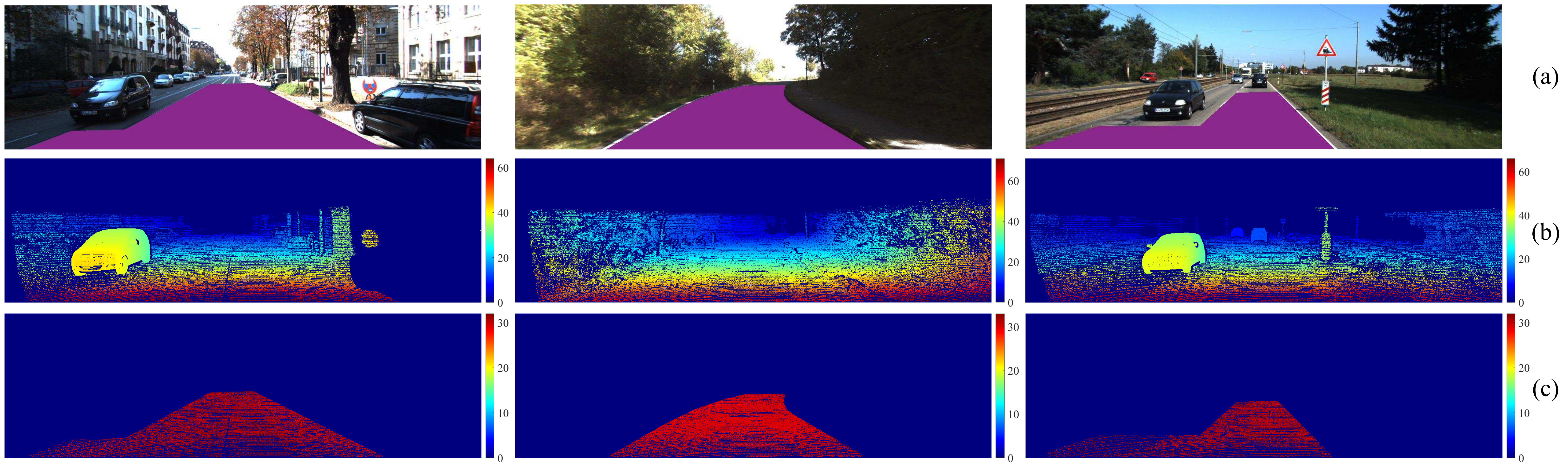}
	\centering
	\caption{Experimental results of the KITTI stereo datasets: (a)  left stereo images (the areas in purple are the labeled road regions);  (b) ground truth disparity maps; (c) transformed disparity maps. }
	\label{fig.kitti_Vision}
\end{figure*}

To evaluate the accuracy of the proposed roll angle estimation algorithm, we utilize a synthesized stereo dataset from EISATS \cite{Vaudrey2008, Wedel2008} where the roll angle is perfectly zero. Some experimental results are shown in Fig. \ref{fig.roll_angle_evaluation}. The road areas (see the magenta regions in the first row of Fig. \ref{fig.roll_angle_evaluation}) are manually selected and the disparities in these areas are utilized to estimate the roll angle $\hat{\theta}$. The absolute difference between the actual and estimated roll angles, i.e.,  $\Delta \theta=|\theta-\hat{\theta}|$, is computed for each frame. The average $\Delta \theta$ is approximately $1.129\times10^{-4}$ rad which is lower than $\frac{\pi}{18000}$ rad. Therefore, the proposed algorithm is capable of estimating the roll angle with high accuracy. 
\subsection{Evaluation of Disparity Transformation}
\label{sec.disp_trans_evaluation}
Since the datasets we created only contain the ground truth of potholes, KITTI stereo datasets \cite{Geiger2012, Menze2015a} are utilized to quantify the performance of our proposed disparity transformation algorithm (the numbers of disparity maps in the KITTI stereo 2012 and 2015 datasets are 194 and 200, respectively). Some experimental results are shown in Fig. \ref{fig.kitti_Vision}. Due to the fact that the proposed algorithm focuses entirely on the road surface, we manually selected a region of interest (see the purple areas in the first row) in each image to evaluate the performance of our algorithm. 
The corresponding transformed disparity maps are shown in the third row of Fig. \ref{fig.kitti_Vision}, where readers can clearly see that the disparities in the road areas tend to have similar values. To quantify the accuracy of the transformed disparity maps, we compute the standard deviation $\sigma_{d}$ of the transformed disparity values as follows: 
\begin{equation}
\begin{split}
\sigma_{d}=\sqrt{\frac{1}{m+1}
	\norm{
		\boldsymbol{\tilde{d}}-\frac{
			\boldsymbol{\tilde{d}}^\top
			\boldsymbol{{1}}_{m+1}
		}{m+1}
	}^2_2}.
\end{split}
\label{eq.sigma}
\end{equation}
where $\boldsymbol{\tilde{d}}=[\tilde{d}_0,\ \tilde{d}_1,\ \cdots,\ \tilde{d}_m]^\top$ is a column vector storing the transformed disparity values. The average $\sigma_d$ values of the two KITTI stereo datasets are provided in Table \ref{table.disp_trans_comparison}. 
\begin{figure}[!t]
	\centering
	\includegraphics[width=0.26\textwidth]{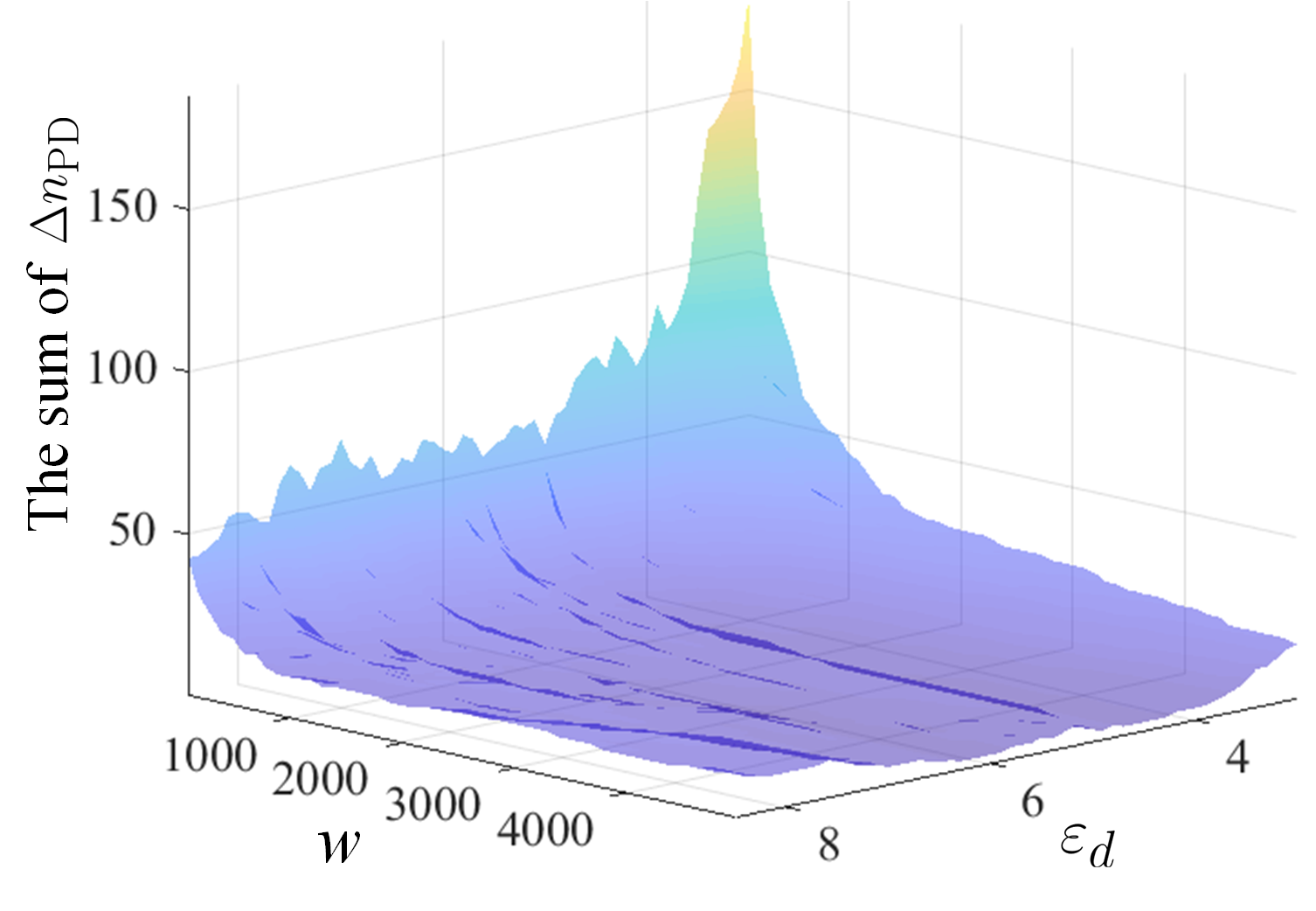}
	\centering
	\caption{The sum of $\Delta n_{\text{PD}}$ with respect to different $\varepsilon_d$ and  $w$.   }
	\label{fig.param_evaluation}
\end{figure}
\begin{table}[!h]
	\begin{center}
		\vspace{0in}
		\footnotesize
		\caption{Comparison of the Average $\sigma_d$ values between the two cases. }
		\label{table.disp_trans_comparison}
		\begin{tabular}{|c|c|c|c|}
			\hline
			Dataset & Case 1 & Case 2 \\
			\hline
			KITTI stereo 2012 training dataset &0.4862 & 0.7800  \\
			KITTI stereo 2015 training dataset & 0.5506 & 0.9428	\\
			\hline
		\end{tabular}
	\end{center}
\end{table}

In case 1, the effects caused by the non-zero roll angle are eliminated, while in case 2, the roll angle is assumed to be zero. 
From Table \ref{table.disp_trans_comparison}, we can observe that the average $\sigma_d$ values of these two datasets reduce by approximately half when the effects caused by the non-zero roll angle are eliminated.  The average $\sigma_d$ value of these two datasets is only 0.5188 pixels.  Therefore, the proposed disparity transformation algorithm performs accurately and the transformed disparity values become very uniform. The runtime of disparity transformation is about 142 ms.  In the next subsection, we will analyze the accuracy of pothole detection.

\subsection{Evaluation of Pothole Detection}
\label{sec.pothole_detection_evaluation}
In Section \ref{sec.pothole_detection_labelling}, a set of randomly selected disparities are modeled as a quadratic surface. The potholes are detected by comparing the difference between the actual disparity map and the modeled quadratic surface. If a connected component contains more than $w$ pixels and the disparity difference of each pixel exceeds $\varepsilon_d$, it will be identified as a pothole. In our experiments, we utilize the brute-force search method to find the best values of $\varepsilon_d$ and $w$. The search range for $\varepsilon_d$ and  $w$ are set to $[3.0, 8.5]$ and $[100,5000]$, respectively. The step sizes for searching $\varepsilon_d$ and $w$ are set to $0.1$ and $100$, respectively. 

\begin{figure}[!t]
	\centering
	\subfigure[]{
		\includegraphics[width=0.213\textwidth]{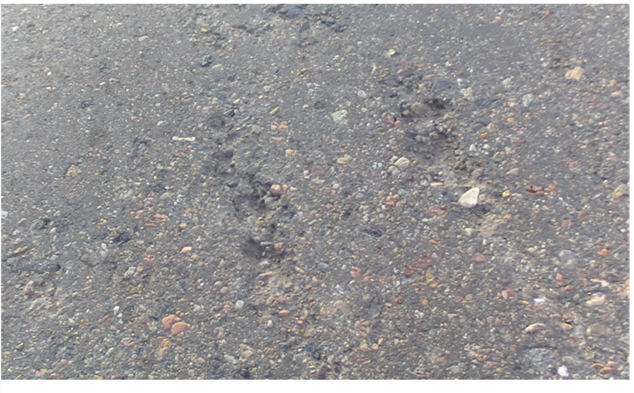}
		\label{fig.failed_left}
	}
	\subfigure[]{
		\includegraphics[width=0.234\textwidth]{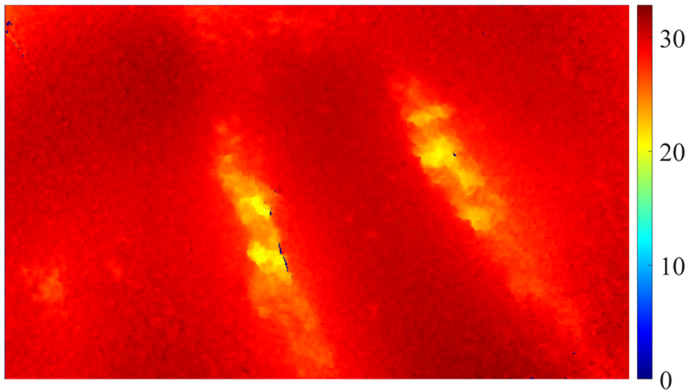}
		\label{fig.failed_disp}
	}
	\subfigure[]{
		\includegraphics[width=0.213\textwidth]{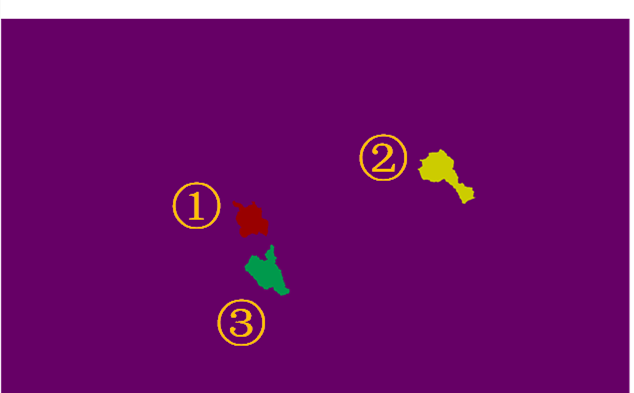}
		\label{fig.failed_result}
	}
	\subfigure[]{
		\includegraphics[width=0.2335\textwidth]{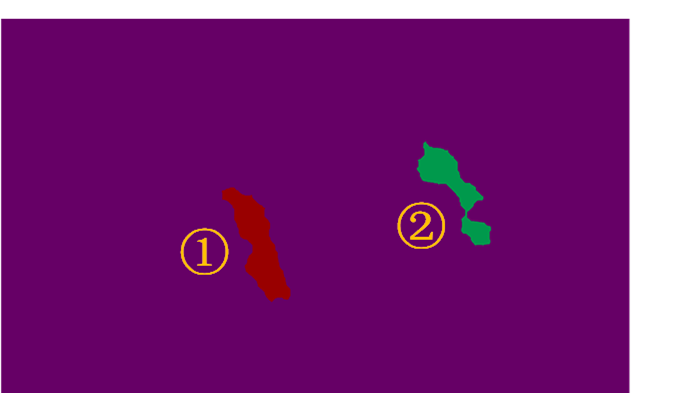}
		\label{fig.failed_gt}
	}
	\caption{Experimental result of incorrect pothole detection: (a) left stereo image; (b) transformed disparity map; (c) detection result; (d) ground truth.   }
	\label{fig.failed_detection}
\end{figure}

\begin{figure*}[!t]
	\centering
	\includegraphics[width=0.82\textwidth]{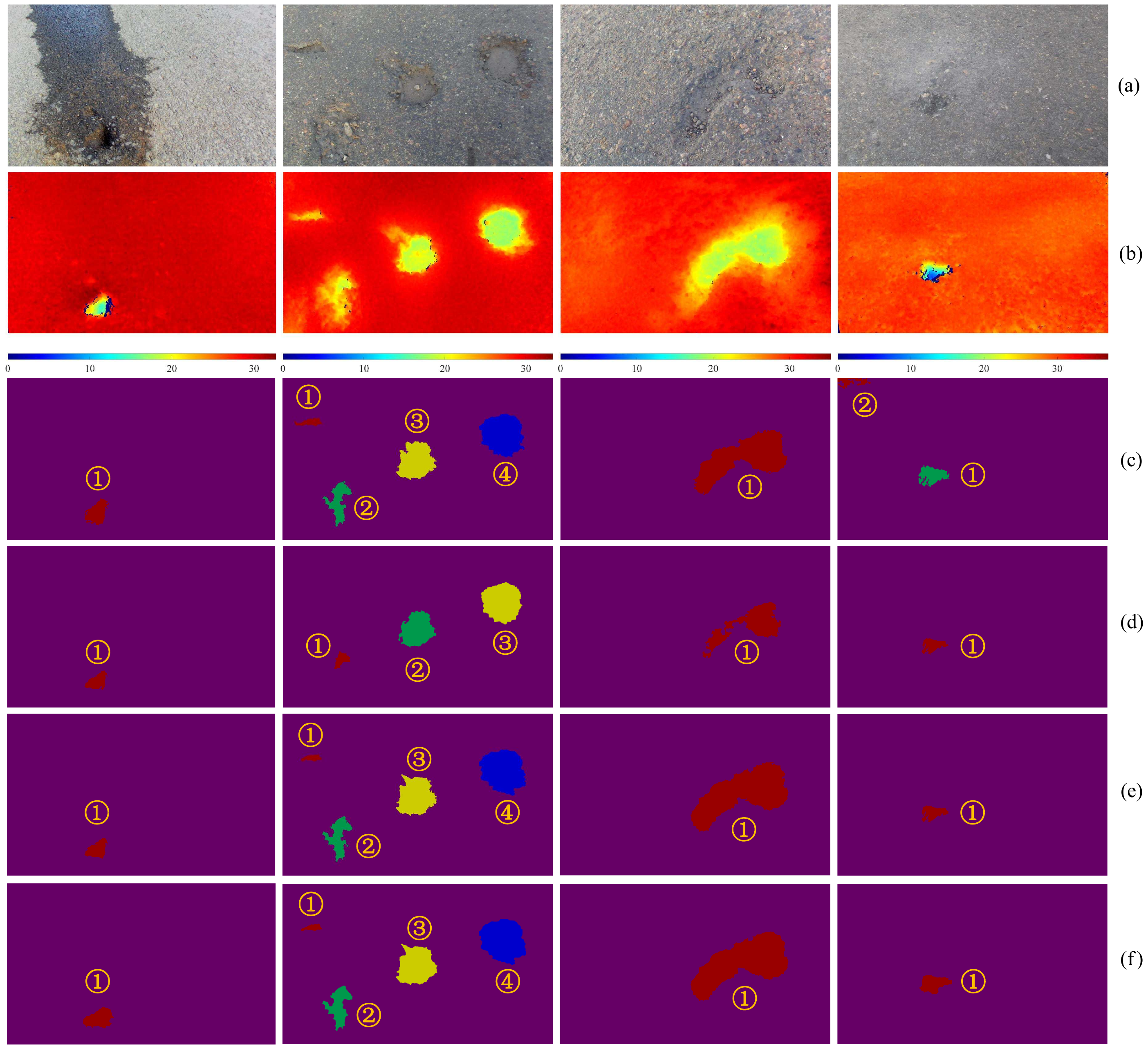}
	\centering
	\caption{Experimental results of pothole detection: (a) left stereo images; (b) transformed disparity maps; (c) pothole detection results obtained using the algorithm proposed in \cite{Zhang2014}; (d) pothole detection results obtained using the algorithm presented in \cite{Mikhailiuk2016}; (e) pothole detection results obtained using the proposed algorithm; (f) pothole ground truth.   }
	\label{fig.detection_results}
\end{figure*}

\begin{table*}[!h]
	\begin{center}
		\vspace{0in}
		\footnotesize
		\caption{Comparison of Successful pothole Detection accuracy. }
		\label{table.successful_detection_rate}
		\begin{tabular}{|c|c|c|c|c|c|c|}
			\hline
			Dataset & Method & \thead{Total\\Potholes} &\thead{Correct\\Detection} & \thead{Incorrect\\Detection} & Misdetection \\
			\hline
			\multirow{3}{*}{Dataset 1} & algorithm in \cite{Zhang2014} &  & 11 & 11 & 0\\
			\cline{4-6}
			& algorithm in \cite{Mikhailiuk2016} & 22 & 22 & 0 & 0 \\
			\cline{4-6}
			& our algorithm &  & 22 & 0 & 0 \\
			\hline
			\multirow{3}{*}{Dataset 2} & algorithm in \cite{Zhang2014} &  & 42 & 10 & 0\\
			\cline{4-6}
			& algorithm in \cite{Mikhailiuk2016} & 52 & 40 & 8 & 4 \\
			\cline{4-6}
			& our algorithm &  & 51 & 1 & 0 \\
			\hline
			\multirow{3}{*}{Dataset 3} & algorithm in \cite{Zhang2014} &  & 5 &0 & 0\\
			\cline{4-6}
			& algorithm in \cite{Mikhailiuk2016} & 5 & 5 &0 & 0 \\
			\cline{4-6}
			& our algorithm &  & 5 &0 & 0 \\
			\hline
			\multirow{3}{*}{Total} & algorithm in \cite{Zhang2014} &  & 58 &21 & 0\\
			\cline{4-6}
			& algorithm in \cite{Mikhailiuk2016} & 79 & 67 &8 & 4 \\
			\cline{4-6}
			& our algorithm &  & 78 &1 & 0 \\
			\hline
		\end{tabular}
	\end{center}
\end{table*}

For the first step, we go through the whole search range and record the number of detected potholes $\hat{n}_\text{PD}$ in each frame. The absolute difference $\Delta n_\text{PD}$ between each $\hat{n}_\text{PD}$ and the expected pothole number  $n_\text{PD}$ is then computed. The sum of $\Delta n_\text{PD}$  with respect to a pair of given $\varepsilon_d$ and  $w$ can therefore be obtained, as illustrated in Fig. \ref{fig.param_evaluation}. In our experiments, the least sum of $\Delta n_\text{PD}$ is equal to one and it is achieved only when $\varepsilon_d=6.2$ and $\varepsilon_d=3100$. The corresponding incorrect pothole detection result is shown in Fig. \ref{fig.failed_detection}. Incorrect detection occurs when the middle of the first pothole subsides and the selected parameters cause the system to detect two potholes instead of one. Some examples of successful detection results are shown in the fifth row of Fig. \ref{fig.detection_results}, and the corresponding ground truth is shown in the sixth row. 

We also compare our proposed algorithm with those produced in \cite{Zhang2014} and \cite{Mikhailiuk2016}. The pothole detection results obtained using the algorithms presented in \cite{Zhang2014} and \cite{Mikhailiuk2016} are shown in the third and forth rows of Fig. \ref{fig.detection_results}, respectively. The comparative pothole  detection results are provided in Table \ref{table.successful_detection_rate}, where we can see that the successful detection accuracy achieved using \cite{Zhang2014} and \cite{Mikhailiuk2016} are $73.4\%$ and $84.8\%$, respectively.  Compared to them, our proposed algorithm can detect potholes more accurately with a successful detection accuracy of $98.7\%$.

We also compare the proposed algorithm with \cite{Zhang2014} and \cite{Mikhailiuk2016} with respect to the pixel-level precision, recall, F-score and accuracy:
\begin{equation}
\text{precision}=\frac{n_\text{TP}}{n_\text{TP}+n_\text{FP}},
\label{eq.precision}
\end{equation}
\begin{equation}
\text{recall}=\frac{n_\text{TP}}{n_\text{TP}+n_\text{FN}},
\label{eq.recall}
\end{equation}
\begin{equation}
\text{F-score}=2\times\frac{\text{precision}\times \text{recall}}{\text{precision}+\text{recall}},
\label{eq.f_score}
\end{equation}
\begin{equation}
\text{accuracy}=\frac{n_\text{TP}+n_\text{TN}}{n_\text{TP}+n_\text{TN}+n_\text{FP}+n_\text{FN}},
\label{eq.accuracy}
\end{equation}
where $n_\text{TP}$, $n_\text{FP}$, $n_\text{FN}$ and $n_\text{TN}$ are true positive, false positive, false negative  and true negative pixel numbers, respectively.

The comparisons with respect to these four indicators are illustrated in Table \ref{table.precision_accuracy_recall}. It can be seen that our proposed algorithm outperforms  \cite{Zhang2014} and \cite{Mikhailiuk2016},  in terms of both pixel-level accuracy and F-score. It achieves an intermediate performance in terms of precision and recall. In addition, the precision and recall achieved using our proposed algorithm are very close to the highest values between \cite{Zhang2014} and \cite{Mikhailiuk2016}. Therefore, the proposed pothole detection algorithm performs both robustly and accurately. 

\begin{table}[!h]
	\begin{center}
		\vspace{0in}
		\footnotesize
		\caption{Comparison of Pixel-Level  Precision, Recall, F-score and  Accuracy.}
		\label{table.precision_accuracy_recall}
		\begin{tabular}{|c|c|c|c|c|c|c|}
			\hline
			Dataset & Method  & recall & precision &  F-score &  accuracy \\
			\hline
			\multirow{3}{*}{Dataset 1} &  \cite{Zhang2014}  & \textbf{0.5199} & 0.5427 & 0.5311 & 0.9892 \\
			\cline{3-6}
			& \cite{Mikhailiuk2016}  & 0.4622 & \textbf{0.9976} & 0.6317 & 0.9936 \\
			\cline{3-6}
			& proposed  & 0.4990 & 0.9871 & \textbf{0.6629} & \textbf{0.9940} \\
			\hline
			\multirow{3}{*}{Dataset 2} &  \cite{Zhang2014}  & 0.9754 & 0.9712 & 0.9733 & 0.9987\\
			\cline{3-6}
			&  \cite{Mikhailiuk2016}  & 0.8736 & \textbf{0.9907} & 0.9285 & 0.9968 \\
			\cline{3-6}
			& proposed  & \textbf{0.9804} & 0.9797 & \textbf{0.9800} & \textbf{0.9991} \\
			\hline
			\multirow{3}{*}{Dataset 3} &  \cite{Zhang2014} & \textbf{0.6119} & 0.7714 & 0.6825 & 0.9948\\
			\cline{3-6}
			&  \cite{Mikhailiuk2016}  & 0.5339 & \textbf{0.9920} & 0.6942 & 0.9957 \\
			\cline{3-6}
			& proposed  & 0.5819 & 0.9829 & \textbf{0.7310} & \textbf{0.9961} \\
			\hline
		    \multirow{3}{*}{Total} &  \cite{Zhang2014} &\textbf{0.7799} & 0.8220 & 0.8004 & 0.9942\\
			\cline{3-6}
			&  \cite{Mikhailiuk2016}  & 0.6948 & \textbf{0.9921} & 0.8173 & 0.9954 \\
			\cline{3-6}
			& proposed  & 0.7709 & 0.9815 & \textbf{0.8635} & \textbf{0.9964} \\
			\hline
		\end{tabular}
	\end{center}
\end{table}

\section{Conclusion and Future Work}
\label{sec.conclusion}
The main contributions of this paper are a novel disparity transformation algorithm and a disparity map modeling algorithm. Using our method, undamaged road areas are better distinguishable in the transformed disparity map and can be easily extracted using Otsu's thresholding method. This greatly improves the robustness of disparity map modeling. To achieve greater processing efficiency, GSS and DP were utilized to estimate the transformation parameters. Furthermore, the disparities, whose normal vectors differ greatly from the optimal one, were also discarded in the process of disparity map modeling, which further improves the accuracy of the modeled disparity map. Finally, the potholes were detected by comparing the difference between the actual and modeled disparity maps. The point clouds of the detected potholes were then extracted from the reconstructed 3D road surface. In addition, we also created three datasets to contribute to stereo vision-based pothole detection research. The experimental results show that the overall successful detection accuracy of our proposed algorithm is around $98.7\%$ and the pixel-level accuracy is approximately $99.6\%$. 

However, the parameters set for pothole detection cannot be applied to all cases. Therefore, we plan to train a deep neural network to detect potholes from the transformed disparity map. Furthermore, road surfaces cannot always be considered to be quadratic. Thus, we aim to design an algorithm to segment the reconstructed road surfaces into a group of localized planes prior to applying the proposed pothole detection algorithm.

\ifCLASSOPTIONcaptionsoff
  \newpage
\fi

\label{sec:refs}
\bibliographystyle{IEEEtran}
\end{document}